\pdfoutput=1

\documentclass[11pt]{article}

\usepackage{EMNLP2023} 

\usepackage{times}
\usepackage{latexsym}
\usepackage{amsfonts}
\usepackage{amsmath}
\usepackage{multirow}
\usepackage{graphicx}
\usepackage{marvosym}
\usepackage{arydshln}
\usepackage{pifont}
\usepackage{fancyhdr}
\usepackage[T1]{fontenc}

\usepackage[utf8]{inputenc}

\usepackage{microtype}

\usepackage{inconsolata}

\usepackage{tabularx}
\usepackage{booktabs}
\usepackage{rotating}
\usepackage{colortbl}
\usepackage{longtable}

\renewcommand\arraystretch{1.15}

\title{UP4LS: User Profile Constructed by Multiple Attributes for Enhancing Linguistic Steganalysis}

\author{Yihao Wang, Ruiqi Song, Lingxiao Li, Yifan Tang, Ru Zhang\textsuperscript{\Letter}, Jianyi Liu \\
\textsuperscript{}School of Cyberspace Security, Beijing University of Posts and Telecommunications, China.\\
\texttt{\{yh-wang, songrq123, lingxiao-li, tyfcs, zhangru, liujy\} @bupt.edu.cn}}

\begin{document}
\maketitle
\begin{abstract}
Linguistic steganalysis (LS) tasks aim to detect whether a text contains secret information. Existing LS methods focus on the deep-learning model design and they achieve excellent results in ideal data. However, they overlook the unique user characteristics, leading to weak performance in social networks. And a few stegos here that further complicate detection. We propose the UP4LS, a framework with the \textbf{U}ser \textbf{P}rofile \textbf{for} enhancing \textbf{LS} in realistic scenarios. Three kinds of user attributes like writing habits are explored to build the profile. For each attribute, the specific feature extraction module is designed. The extracted features are mapped to high-dimensional user features via the deep-learning model of the method to be improved. The content feature is extracted by the language model. Then user and content features are integrated. Existing methods can improve LS results by adding the UP4LS framework without changing their deep-learning models. Experiments show that UP4LS can significantly enhance the performance of LS-task baselines in realistic scenarios, with the overall Acc increased by 25\%, F1 increased by 51\%, and SOTA results. The improvement is especially pronounced in fewer stegos. Additionally, UP4LS also sets the stage for the related-task SOTA methods to efficient LS.
\end{abstract}


\section{Introduction}\label{sec1}
Linguistic steganography is an information concealment technique that involves embedding secrets within texts and transmitting these texts through an open channel \cite{ADG}. Only authorized recipients can perceive the existence of the stegos and extract secrets. This technology leads to slight differences in distributions compared to ``covers'' (natural texts)  \cite{RNN-Stega}\cite{GAN-APD}. Linguistic steganalysis (\textbf{LS}) tasks aim to extract such slight differences to determine whether texts are ``stegos'' (texts generated by steganography). Two types of LS have been proposed: manual construction \cite{traditional-ls} and automatic extraction \cite{Few-shot}\cite{LSFLS}. The former focuses on the development of effective manual features, such as word associations \cite{n-gram+SVM}, which are interpretable and targeted for extraction. These features are specifically extracted to capture the differences between covers and stegos, and it has good results on the specific LS tasks. The latter employs deep-learning models to extract high-dimensional features. These features have a robust capacity to quantify steganographic embedding, resulting in superior performance on the broad LS tasks. Therefore, in recent years researchers have focused on this type of LS.

\begin{figure}[!b]
	\centering
	\includegraphics[width=0.47\textwidth]{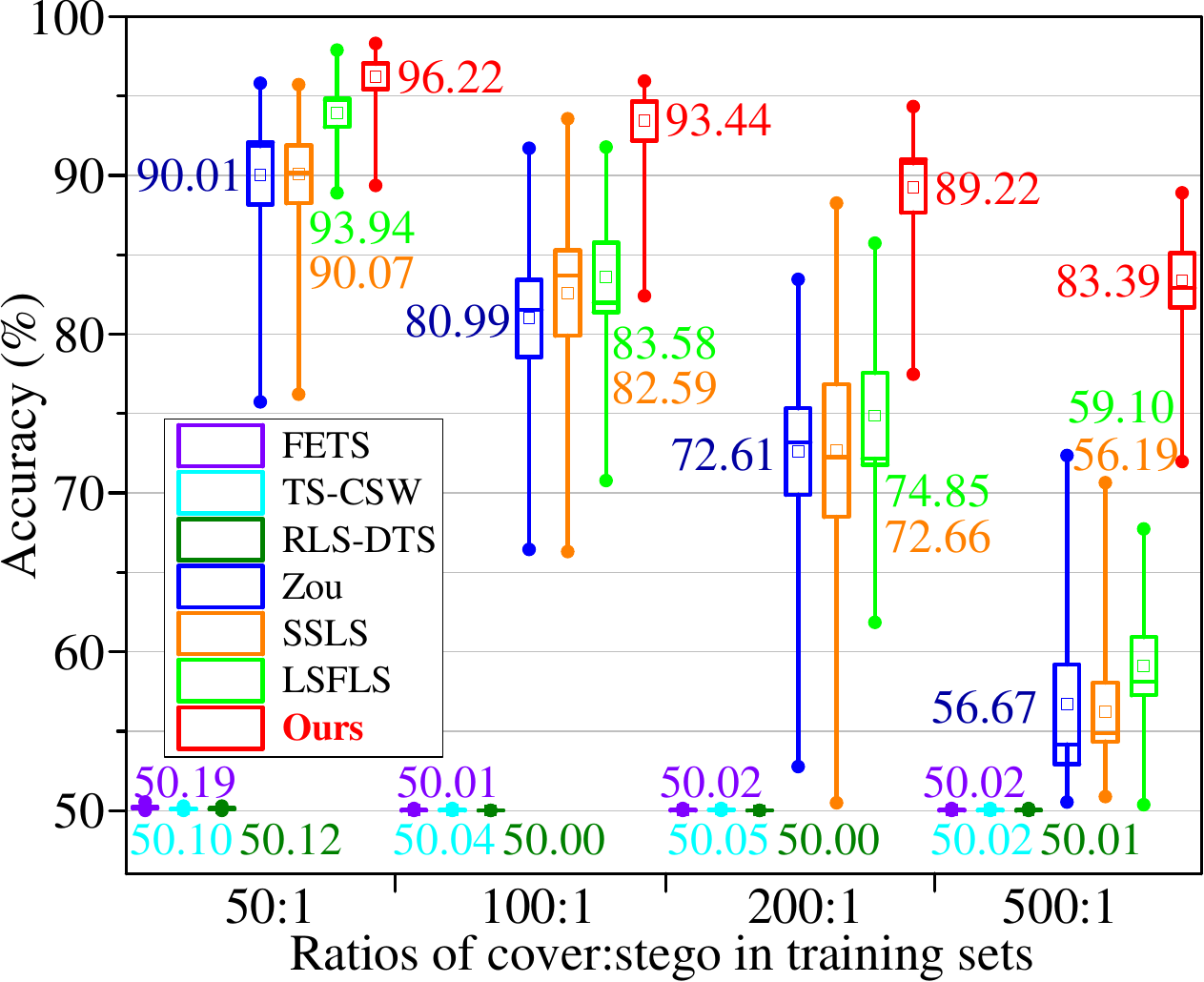}
	\caption{Detection results of LS methods in datasets with various ratios (cover:stego). The box plot depicts the overall performance on 10 user data, as introduced in Section \ref{sec31}. In each box, the hollow squares are the average value in 10 values, as marked by the labels.}
	\label{intro}
	\vspace{-0.6ex}
\end{figure}

Recent LS work has been proposed with novel motivations. To improve the performance of ideal stegos, Zou et al. \cite{Zou} extracted global features and captured the critical part among them, greatly improving the performance. To effectively detect stegos in few-shot scenarios, Wang et al. \cite{LSFLS} and Wen et al. \cite{Few-shot} designed methods to achieve excellent performance. Due to the different domains, ordinary methods find it difficult to detect stego in cross-domain data. Xue et al. \cite{mda} and Wang et al. \cite{RLS-DTS} successively proposed cross-domain LS based on domain adaptation and reinforcement learning, and achieved excellent performance on cross-domain datasets.

Social networks are regarded as one of the primary channels for transmitting stegos. Due to their convenience and diverse applications, they have gained immense popularity, hence the demand for LS within this environment has surged. To evaluate the detection effectiveness of existing LS in social networks, we utilize six prevailing LS methods: FETS \cite{FEFT}, TS-CSW \cite{TS_CSW}, RLS-DTS \cite{RLS-DTS}, Zou \cite{Zou}, SSLS \cite{SSLS}, and LSFLS \cite{LSFLS}. The datasets consist of covers posted by Twitter users and stegos generated by the ADG (Adaptive Dynamic Grouping) algorithm \cite{ADG}. This algorithm is known for its strong concealment capabilities in both theory and practice. To simulate the real social network as much as possible, the quantity of stego is smaller than that of cover. We varied the ratios of cover:stego from 50:1 to 500:1 in the training sets, while ensuring a uniform ratio of 1:1 in the testing sets. Further details about the experimental settings can be found in Section \ref{sec31}. Figure~\ref{intro} illustrates the detection performance of existing LS methods in datasets with various ratios.

The results in Figure~\ref{intro} show that the performance of the existing methods has insufficient performance for a small number of stego in social network scenarios, and the performance drops notably as the ratio increases. This phenomenon is because social network posts exhibit unique user characteristics influenced by various user attributes, resulting in strong user personalization. These user characteristics are difficult to imitate in stegos. However, existing LS methods ignore users' personalized characteristics, resulting in limited effective detection in social networks. Moreover, compared to the vast quantity of covers in social networks, the quantity of stegos is exceedingly small, which poses a substantial challenge for detection.

In this work, we propose the UP4LS, a novel framework with the \textbf{U}ser \textbf{P}rofile \textbf{for} enhancing the \textbf{LS} performance of existing methods. UP4LS leverages the potential user attributes reflected in post, thereby creating user profiles. Then we designed a targeted feature extraction module for each user attribute, and the extracted features will be mapped to high-dimensional user features. The content feature is also extracted. It guides and learns by user features, and the two types of features are concatenated, further improving feature representation. UP4LS increases sensitivity to stegos during training. To facilitate the transplantation of existing methods, the deep-learning model in existing methods are retained. The remaining components are modified according to UP4LS, which can be used for steganalysis in social networks. Experiments show that UP4LS not only improves the performances of prevailing LS-task baselines, but also provides a platform for related-task SOTA methods to conduct effective LS.

Our main contributions are outlined below.

\begin{itemize}
	\item To improve LS in social networks, UP4LS innovatively built the user profile for LS. The attributes of the user profile are derived from posts, and they are habits, psychology, and focus. Specific feature extraction is designed for every user attribute to extract user features.
\item To improve feature representation, we employ the attention mechanism to guide the learning of content features by user features. Then they are concatenated to obtain the LS feature.
\item To evaluate UP4LS performance, we collect posts from multiple users and generate stegos with various ratios. Results show that UP4LS not only improves the performance of LS-task baselines but also opens new avenues for related-task SOTA methods on LS tasks.
\end{itemize}


\begin{figure*}[!htbp]
	\centering
	\includegraphics[width=6.3in]{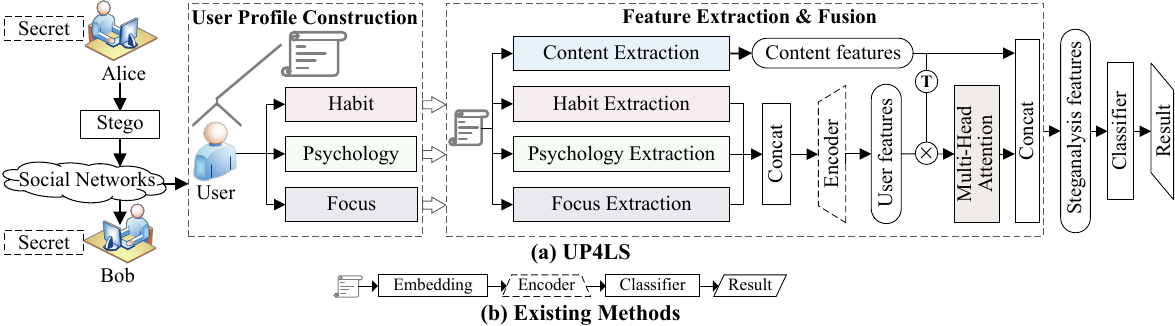}
	\caption{The overall architecture of UP4LS. UP4LS consists of two modules: ``\textbf{User Profile Construction}'' and ``\textbf{Feature Extraction \& Fusion}''. ``\textbf{(b) Existing Methods}'' provides the overall architecture of existing methods. UP4LS takes in texts as input, a mixture of covers and stegos. This user profile is divided into three types of user attributes: ``\textbf{Habit}'', ``\textbf{Psychology}'', and ``\textbf{Focus}''. To enhance the performance of existing methods, they only need to retain the ``Encoder'' component, and the rest is modified according to UP4LS.}
	\label{up4ls}
	\vspace{-2.5ex}
\end{figure*}

\section{Related Work}\label{sec4}
\paragraph {Generative linguistic steganography.} Linguistic steganography aims to automatically generate stego texts that have secret information \cite{RNN-Stega}. Fang et al. \cite{Fang} construct a linguistic steganography system, which is capable of generating high-quality stegos. Yang et al. \cite{RNN-Stega} design two text-coding methods based on conditional probability distributions to generate stegos. Zhang et al. \cite{ADG} establish a dynamic encoding method for embedding secret information, which adaptively and dynamically groups tokens and embeds them using the probabilistic recurrence given by the language model. Wang et al. \cite{LLsM} propose the LLsM, which is the steganography work based on open-source and closed-source LLMs.

\paragraph {Linguistic steganalysis.} To prevent criminals from using generative linguistic steganography to transmit secret information, LS has been developing in recent years. LS can effectively detect generative stego texts, which are confirmed by a series of representative works. Because a single LSTM module makes it difficult to extract enough low-level features, Yang et al. \cite{bld} present a method to densely connect LSTM based on feature pyramids. Wu et al. \cite{GCN} apply GNN for LS. This method transforms text into a directed graph that has relevant information. Yang et al. \cite{sesy} design a novel framework to keep and make full use of the syntactic structure by integrating semantic and syntactic features of the texts. Xue et al. \cite{mda}\cite{insa} devote to a domain-adaptive steganalysis method and an alternative hierarchical mutual-learning LS framework. These methods separately resolve the problem that the scale of the model is too large and the problem that performance is low due to domain mismatch.

Table \ref{rw} shows the overview of LS works. It involves whether BERT-based, the architecture, whether the covers come from social users, and if the quantity of cover and stego is unbalanced.

\begin{table}[!htbp]
	\centering
	{\fontsize{7.8pt}{8.8pt}\selectfont
		\setlength{\tabcolsep}{0.5mm}
		\caption{The overview of LS works.}\label{rw}
		\vspace{-2ex}
		\begin{tabular}{ccccc}
			\hline
			\textbf{References} & \textbf{BERT}&\textbf{Architecture} &\textbf{From users} & \textbf{Unbalanced}\\
			\hline
			\cite{FEFT}		&\ding{55}	&FCN		&\ding{55}	&\ding{55}\\
			\cite{TS_BiRNN}	&\ding{55}	&RNN		&\ding{55}	&\ding{55}	\\
			\cite{bld}&\ding{55}	&LSTM		&\ding{55}	&\ding{55}	\\	
			\cite{GCN}	&\ding{55}	&GNN		&\ding{55}	&\ding{55}	\\
			\cite{Zou}			&\ding{51}	&LSTM		&\ding{55}	&\ding{55}	\\
			\cite{SSLS}		&\ding{51}	&GRU, CNN	&\ding{55}	&\ding{55}	\\
			\cite{mda}			&\ding{51}	&CNN		&\ding{55}	&\ding{55}	\\
			\cite{Few-shot}&\ding{51}	&LSTM		&\ding{55}	&\ding{55}	\\
			\cite{sesy}		&\ding{51}	&GAT		&\ding{55}	&\ding{55}	\\
			\cite{RLS-DTS}&\ding{55}	&Actor-Critic	&\ding{55}	&\ding{55}	\\
			\cite{EILG}		&\ding{55}	&GRU, CNN	&\ding{55}	&\ding{55}	\\
			\cite{LSFLS}		&\ding{51}	&BNN		&\ding{55}	&\ding{51}	\\	
			\hline
			\textbf{Ours (UP4LS)}&\ding{51}	&User Profile	&\ding{51}	&\ding{51}	\\
			\hline
		\end{tabular}
	}
	\vspace{-1ex}
\end{table}

\section{Methodology}\label{sec2}


\subsection{UP4LS Overall}\label{sec21}
Under the existing ideal experimental environments, almost all LS methods are focused on capturing content features like semantics and grammar \cite{TS_CSW}\cite{SSLS}\cite{2023ACMTrans}. However, these methods usually overlook the subjective aspects of human expression in writing. As a result, the LS effectiveness tends to be suboptimal when applied to social networks. Therefore, we propose the UP4LS framework, which improves the performance of existing methods for LS in social networks. Figure~\ref{up4ls} illustrates the overall architecture of UP4LS.
\subsection{User Profile Construction}\label{sec22}

\begin{figure}[!b]
	\centering
	\includegraphics[width=0.44\textwidth]{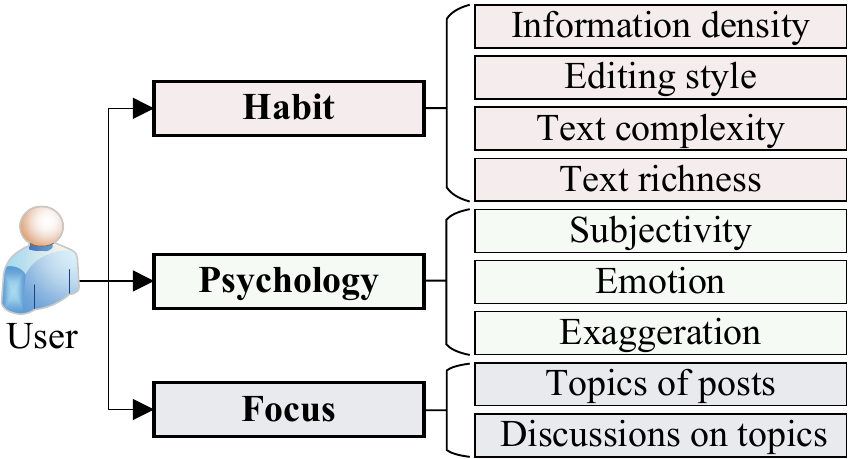}
	\caption{The specific user profile for LS.}
	\label{up}
	\vspace{-0.7ex}
\end{figure}

\begin{figure*}[!h]
	\centering
	\includegraphics[width=5.2in]{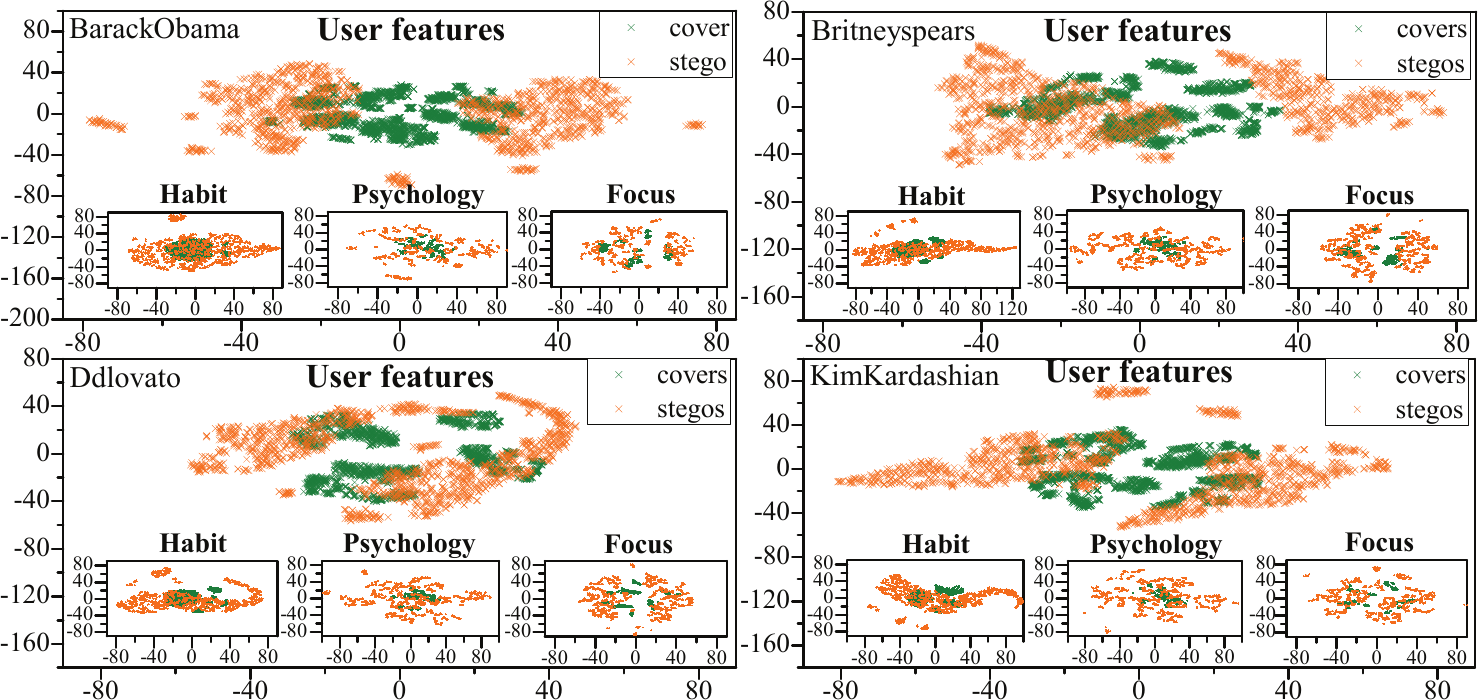}
	\caption{Distribution of covers and stegos in user feature space extracted by UP4LS. Taking 4 users as examples, their usernames are presented in the upper left corner. For more details about the user datasets in Section \ref{sec31}. We use t-SNE \cite{t-SNE} to visualize the user features of texts. The green and orange marks represent the feature distribution of covers and stegos. Each subfigure contains three small figures, which are the feature distribution of ``\textbf{Habit}'', ``\textbf{Psychology}'', and ``\textbf{Focus}''. These user features in this figure are not backpropagated, they are directly extracted in one go. This figure serves to show the rationality and effectiveness of user features for LS tasks.}
	\label{Distribution}
	\vspace{-2.5ex}
\end{figure*}

\paragraph {User Profile for LS.} From a macro perspective, the construction of the general user profile can effectively improve decision-making effects by analyzing user characteristics and behaviors \cite{UP1}\cite{UP2}. Currently, there is no steganography that can combine content and user behavior \cite{behavior} for information hiding. So we focus on the content of user posts itself. Figure~\ref{up} illustrates the user profile for LS.

\paragraph {Habit.} It involves ``information density'', ``editing style'', ``text richness'', and ``text complexity''. Users exhibit a unique writing style within their posts. This uniqueness often stems from the user's growth background, cultural upbringing, and life experience. Each user's distinctive upbringing adds personalization to the expression.

\paragraph {Psychology.} It involves ``subjectivity'', ``emotion'', and ``exaggeration''. Subjectivity in a post can reveal a user’s opinion tendencies. Some users may display strong subjectivity when expressing their opinions, while some users may prioritize objective facts. The degree of exaggeration embodied in a post can reveal a user's specific style. Analyzing psychology helps obtain personalized characteristics such as long-term and short-term emotional dispositions.

\paragraph {Focus.} It involves ``topics of posts'' and ``discussions on topics''. Users' areas of focus often reflect their knowledge and interests. This selective focus can indicate their social role, professional background, or current life stage.


\subsection{Feature Extraction \& Fusion}\label{sec23}
\paragraph {User Features for LS.} Current steganography struggles to imitate user characteristics, which results in differences between covers and stegos in this dimension. Capturing these differences and extracting such features can improve LS.

To better capture these differences, we designed a feature extraction module for each user attribute within the user profile. These modules include ``Habit Extraction'', ``Psychology Extraction'', and ``Focus Extraction''. Figure~\ref{Distribution} illustrates the distribution of covers and stegos in user feature space, and this figure explains that user features are reasonable and effective for LS tasks.

\paragraph {Habit Extraction.} This is the first module for these extraction modules. It aims to capture various aspects of writing habits, encompassing factors like ``Information density'', ``Editing style'', ``Text richness'', and ``Text complexity''. Users usually reflect their underlying writing habits when editing posts, and it is difficult for existing steganography to completely imitate these habits.

``Information density'' is captured by analyzing the scale and distribution of nouns, pronouns, and verbs within the text.

``Editing style'' is determined by examining the scale and distribution of function words \cite{Editor1}\cite{Editor2}\cite{Editor3}, such as prepositions, determiners, and coordinating conjunctions.

``Text richness'' is evaluated by capturing the scale and distribution of adjectives and adverbs. To perform this analysis, NLTK\footnote[1]{https://www.nltk.org/} is used for part-of-speech tagging, enabling us to count the scale and distribution of various words based on the tagging.

\begin{figure}[!htbp]
	\centering
	\includegraphics[width=0.46\textwidth]{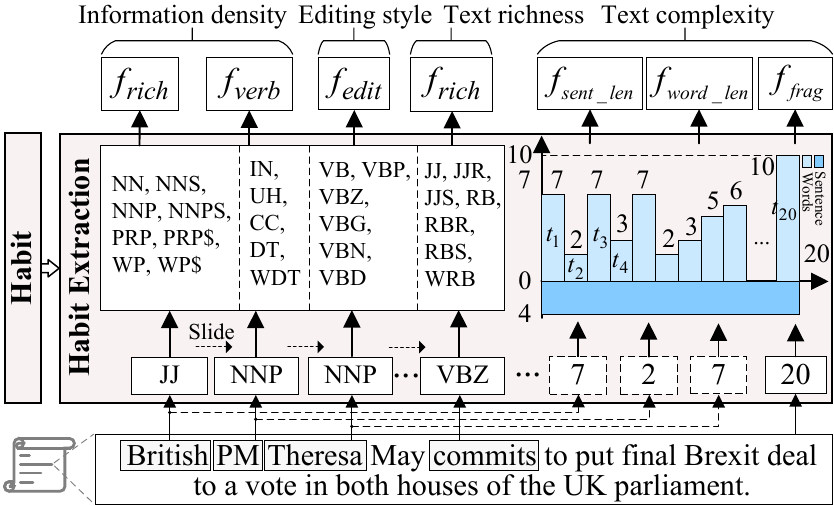}
	\caption{The working principle of the ``Habit Extraction''. The input of this module is text, and the output is extracted features about the dimension of ``Information density'', ``Editing style'', ``Text richness'', and ``Text complexity''.}
	\label{Habit}
\end{figure}

``Text complexity'' is quantified by calculating sentence length, word length, and scale and distribution of symbols. Typically, spoken texts exhibit simplified grammar, shorter sentences, and shorter word lengths. Increased usage of punctuation marks within a sentence indicates more pauses, leading to a higher degree of fragmentation and a stronger oral language nature. Conversely, a more pronounced written style features a reduced frequency of punctuation marks, there is ${f_{frag}} = {1 \mathord{\left/ {\vphantom {1 {count(\mathrm{punc})}}} \right. \kern-\nulldelimiterspace} {count(\mathrm{punc})}},\mathrm{punc} = \{ ,.;?! \cdots \}$ . Figure~\ref{Habit} illustrates the working principle of the ``Habit Extraction''.

\vspace{-3ex}

\paragraph {Psychology Extraction.} It is the second module for these extraction modules. To analyze ``Subjectivity'' and ``Emotion'', TextBlob\footnote[2]{https://textblob.readthedocs.io/en/dev/} library is employed to provide a set of APIs that simplify common text analysis tasks. In recent years, TextBlob has gained significant attention for its outstanding performance in sentiment analysis \cite{TextBlob1}\cite{TextBlob2}. During emotional calculations, TextBlob uses a dictionary that encompasses parameters like ``polarity'', ``subjectivity'', and ``intensity''. Given a text input, it returns a named tuple representing sentiment and subjectivity as ``(polarity, subjectivity)''. The formulas are shown below.

\vspace{-3ex}
\begin{equation}
\small
\mathrm{Emotion} = \frac{{\sum\limits_{i = 0}^K {{{( - 0.5)}^n} \times {S_{i\_adverb}} \times {S_{punc}}} }}{{K/{S_{emoticon}}}},
\end{equation}
\vspace{-3ex}
\begin{equation}
\small
{S_{i\_adverb}} = {\mathrm{max}}( - 1,\mathrm{min}({S_i} \times {S_{adverb}},1)),
\end{equation}
\vspace{-4ex}
\begin{equation}
\small
\mathrm{Subjectivity} = {\mathrm{max(0}},\mathrm{min}(\sum\limits_{i = 0}^K {S{'_i} \times S{'_{adverb}}} ,1)),
\end{equation}
\vspace{-4ex}

\noindent where, $K$ is the number of words related to emotional polarity and subjectivity in the text. ${S_{i\_adverb}},{S_{punc}},$ and ${S_{emoticon}}$ represent the emotional value of adverbs, punctuation, and expressions of various degrees. $S{'_i}$ and $S{'_{adverb}}$ represent the subjective value of the current emotional word and emotional adverb. $n$ represents the number of negative words related to the current emotional vocabulary. The ``Exaggeration'' features are captured by analyzing the frequency of interjections. 

Consider that users may have different habits when expressing emotions, resulting in varying degrees of exaggeration in text. The interjection is a significant feature \cite{interjections1}\cite{interjections2}. We define interjections as words that are longer than four letters but have fewer than half the number of unique letters in total length. The formula is shown below.

\vspace{-3ex}
\begin{equation}
\small
{f_{exag}} = \left\{ {\begin{array}{*{20}{c}}
	{0,{\mathrm{  }}else} \\ 
	{\frac{1}{{c({t_i})}},{\mathrm{  }}len({t_i}) > 4\& c(t_i^r) \geq \frac{len({t_i})}{2}} 
	\end{array}} \right.,
\end{equation}

\noindent where, $c( \cdot )$ is the count, and $t_i^r$ is the repeated character ${t_i}$.

\paragraph {Focus Extraction.} It is the last module for these extraction modules. We employ Latent Dirichlet Allocation (LDA) \cite{LDA} to analyze the ``topics of posts''. Given a collection of document $\mathbb{D} = \{ {D_1},{D_2}, \cdots, {D_j}\}$ and a predefined number of topics, denoted as $k$. 

In social networks, users often include hyperlinks when ``discussing on topics''. These hyperlinks, typically consisting of irregular character strings, are unlikely to be present in the vocab. Stego is generated based on the vocab, the probability of a hyperlink appearing in it is very low.

\paragraph {Encoder.} Existing LS methods focus on the design of ``Encoder'', such as LSTM-based \cite{Zou} and CNN-based \cite{SSLS}. They achieved excellent detection performance in ideal data. To improve their detection performance in social network scenarios, in UP4LS, their respective Encoder architectures will be retained, and other modules can be modified to the UP4LS design to improve their performance in this realistic scenario.

\paragraph {Content Features.} BERT \cite{BERT} is employed to extract content feature. It is not this paper's focus, so we do not introduce it in detail.

\paragraph {Feature Fusion.} Since user features ${F_{user}}$ and ${F_{content}}$ are not the same dimension, direct concatenating may result in insufficient performance. We use the mutual attention to interact with them. The attention matrix $Attn$ is obtained. UP4LS then concatenates $Attn$ and ${F_{content}}$ to get the final LS features $F$. The formulas are shown below.

\vspace{-2ex}
\begin{equation}
F = {\text{Concat}}(Attn,{F_{content}}),
\end{equation}
\vspace{-4ex}
\begin{equation}
Attn = \frac{{Q \times {K^{\text{T}}}}}{{\sqrt {{d_k}} }} = \frac{{{F_{user}} \times {F_{content}}^{\text{T}}}}{{\sqrt {{d_{{F_{content}}}}} }},
\end{equation}

\noindent where, ${d_{{F_{content}}}}$ is the dimension of ${F_{content}}$ and T is the transpose operation.

\subsection{Training}\label{sec24}
During the training phase, we optimizes commonly used cross-entropy loss of LS work, making training more focused on stego samples. The formulas of the loss functions are shown below.

\begin{equation}
\begin{aligned}
\mathcal{L}_{p_t} =& - \alpha_t \big[ (1 - p_t)^{\gamma + 1} \log (1 - p_t) \\
& + p_t (1 - p_t)^\gamma \log (p_t) \big]
\end{aligned}
\end{equation}
\noindent where, $\gamma$ is the adjustment factor, $p_t$ is the probability, and ${\alpha_t}$ is the loss weight of the stego.


\section{Experiments}\label{sec3}

To ensure fairness and reliability in comparisons between methods, each experiment was repeated 5 times for every dataset, and the results were averaged to provide the results. Experiments are run on the NVIDIA GeForce RTX 3090 GPU.


\subsection{Settings}\label{sec31}
\paragraph {Dataset.} We constructed datasets with four ratios of cover:stego. The ratios are 50:1, 100:1, 200:1, and 500:1 in the training sets. The ratio is 1:1 in testing sets. Datasets are divided into training, validation, and testing sets of 6:2:2. In each dataset, covers come from posts by 10 users. Stegos are generated by the high-performance steganography ADG \cite{ADG}. ADG security has been analyzed through proof and practice. Table~\ref{Dataset} shows the specific information of the dataset.

\begin{table}[!ht]
	\centering
	\small
	\setlength{\tabcolsep}{1mm}
	\caption{The specific information of the datasets. (Take the num of stegos as 200:1 as an example. ``ER'' represents the embedding rate of the stegos)}\label{Dataset}
	\begin{tabular}{ccccccc}
		\toprule[1pt]
		\multirow{2}*{No.} & \multirow{2}*{Name} & \multicolumn{2}{c}{Training} &\multicolumn{2}{c}{Testing} & \multirow{2}*{ER} \\
		~ & ~ & covers & stegos&  covers & stegos & ~ \\ \midrule[0.6pt]
		U1 & ArianaGrande 	& 2,325 & 11 	& \multicolumn{2}{c}{580} 	& 3.88 \\ 
		U2 & BarackObama 	& 2,291 &	11	& \multicolumn{2}{c}{572}	& 4.20 \\ 
		U3 & BritneySpears 	& 2,194 & 10	& \multicolumn{2}{c}{548}	& 5.06 \\ 
		U4 & Cristiano 		& 1,940 &	9	& \multicolumn{2}{c}{485}& 4.54 \\ 
		U5 & Ddlovato 		& 1,703 &	8	& \multicolumn{2}{c}{425}& 4.78 \\ 
		U6 & JimmyFallon 	& 2,455 &	12	& \multicolumn{2}{c}{613}& 3.91 \\ 
		U7 & Justinbieber 	& 1,660 &	8	& \multicolumn{2}{c}{414}& 4.12 \\ 
		U8 & KimKardashian 	& 2,351 &	11	& \multicolumn{2}{c}{587}& 4.85 \\ 
		U9 & Ladygaga 		& 1,840 &	9	& \multicolumn{2}{c}{459}& 5.18 \\ 
		U10 & Selenagomez 	& 2,243 &	11	& \multicolumn{2}{c}{560}& 4.39 \\ \bottomrule[1pt]
	\end{tabular}
\end{table}

\paragraph {Baselines.} The baselines consist of two parts, that is \underline{LS-task} and \underline{related-task} baselines.

The \underline{LS-task} baselines include: 

non-BERT-based: \textbf{1. FETS} \textsubscript{(\textit{IEEE SPL})} \cite{FEFT}, which has shown superior performance compared to manual constructive methods, and \textbf{2. TS\_RNN} \textsubscript{(\textit{IEEE SPL})} \cite{TS_BiRNN}, which exhibits excellent performance on multiple ideal datasets. BERT-based: \textbf{3. Zou} \textsubscript{(\textit{IWDW})} \cite{Zou}, which achieved high performance, \textbf{4. SSLS} \textsubscript{(\textit{IEEE SPL})} \cite{SSLS}, which displays remarkable performance on mixed sample sets, and \textbf{5. LSFLS} \textsubscript{(\textit{IEEE TIFS})} \cite{LSFLS}, which achieves SOTA performance in the few-shot data.

The \underline{related-task} baselines include: 

Fine-grained emotion classification tasks: \textbf{6. HypEmo} \textsubscript{(\textit{ACL})} \cite{HypEmo}, which employs hyperbolic space to capture hierarchical structures. It performs SOTA when the label structure is complex or the relationship between classes is ambiguous. Hierarchical text classification tasks: \textbf{7. HiTIN} \textsubscript{(\textit{ACL})} \cite{HiTIN}, which uses a tree isomorphism network to encode the label hierarchy. It performs well in large-scale hierarchical tasks.

Given these methods' widely recognized performance on specific tasks.

\begin{table*}[!htbp]
	\centering
	{\fontsize{8.5pt}{10pt}\selectfont
		\setlength{\tabcolsep}{0.28mm}
		\caption{Overall comparison of the original \textbf{LS-task} baselines (Original) and with UP4LS (+UP4LS) in the distinct datasets. ``a\textsubscript{$\pm$b \textcolor{red}{(c)}}'' represents ``Average\textsubscript{$\pm$Standard Deviation \textcolor{red}{($\Delta$Acc)}}''. $\Delta$ is +UP4LS $-$ Original, indicated by \textcolor{red}{Red} value. \textbf{Bold} value represents the best performance. The Unit is \%. The complete data are shown in Table \ref{a-1} to Table \ref{a-4} in Appendix \ref{appendix}.}\label{ls-task}
		\begin{tabular}{cccccccccc}
			\toprule[1pt]
			\multicolumn{2}{c}{\multirow{2}*{\textbf{LS-task (\%)}}}  & \multicolumn{2}{c}{50:1}   & \multicolumn{2}{c}{100:1}     & \multicolumn{2}{c}{200:1}     & \multicolumn{2}{c}{500:1} \\
			&       & Original & +UP4LS & Original & +UP4LS & Original & +UP4LS & Original & +UP4LS \\
			\midrule[0.6pt]
			
			\multirow{2}*{FETS} & Acc   & 50.19\textsubscript{$\pm$0.20} & \textbf{95.59}\textsubscript{$\pm$3.12 \textcolor{red}{($\uparrow$45.40)}} & 50.01\textsubscript{$\pm$0.04} & \textbf{93.08}\textsubscript{$\pm$4.25 \textcolor{red}{($\uparrow$43.07)}} & 50.01\textsubscript{$\pm$0.04} & \textbf{88.81}\textsubscript{$\pm$5.52 \textcolor{red}{($\uparrow$38.80)}} & 50.01\textsubscript{$\pm$0.04} & \textbf{82.75}\textsubscript{$\pm$7.27 \textcolor{red}{($\uparrow$32.74)}} \\
			& F1    & 0.73\textsubscript{$\pm$0.78} & \textbf{95.35}\textsubscript{$\pm$3.66 \textcolor{red}{($\uparrow$94.62)}} & 0.05\textsubscript{$\pm$0.15} & \textbf{92.56}\textsubscript{$\pm$5.20 \textcolor{red}{($\uparrow$92.51)}} & 0.05\textsubscript{$\pm$0.15} & \textbf{86.96}\textsubscript{$\pm$8.47 \textcolor{red}{($\uparrow$86.91)}} & 0.05\textsubscript{$\pm$0.15} & \textbf{78.36}\textsubscript{$\pm$11.33 \textcolor{red}{($\uparrow$78.31)}} \\
			\midrule[0.6pt]
			
			\multirow{2}*{TS\_RNN} & Acc   & 50.11\textsubscript{$\pm$0.15} & \textbf{96.76}\textsubscript{$\pm$2.00 \textcolor{red}{($\uparrow$46.65)}} & 50.05\textsubscript{$\pm$0.07} & \textbf{93.30}\textsubscript{$\pm$4.10 \textcolor{red}{($\uparrow$43.25)}} & 50.02\textsubscript{$\pm$0.04} & \textbf{88.66}\textsubscript{$\pm$4.67 \textcolor{red}{($\uparrow$38.64)}} & 50.01\textsubscript{$\pm$0.04} & \textbf{83.27}\textsubscript{$\pm$5.97 \textcolor{red}{($\uparrow$33.26)}} \\
			& F1    & 0.44\textsubscript{$\pm$0.61} & \textbf{96.67}\textsubscript{$\pm$2.11 \textcolor{red}{($\uparrow$96.23)}} & 0.20\textsubscript{$\pm$0.27} & \textbf{92.88}\textsubscript{$\pm$4.62 \textcolor{red}{($\uparrow$92.68)}} & 0.08\textsubscript{$\pm$0.18} & \textbf{86.89}\textsubscript{$\pm$6.23 \textcolor{red}{($\uparrow$86.81)}} & 0.05\textsubscript{$\pm$0.15} & \textbf{79.60}\textsubscript{$\pm$9.70 \textcolor{red}{($\uparrow$79.55)}} \\
			\midrule[0.6pt]
			
			\multirow{2}*{Zou} & Acc   & 90.01\textsubscript{$\pm$5.82} & \textbf{95.95}\textsubscript{$\pm$2.79 \textcolor{red}{($\uparrow$5.94)}} & 80.99\textsubscript{$\pm$7.72} & \textbf{93.91}\textsubscript{$\pm$3.71 \textcolor{red}{($\uparrow$12.92)}} & 72.61\textsubscript{$\pm$8.61} & \textbf{89.44}\textsubscript{$\pm$4.95 \textcolor{red}{($\uparrow$16.83)}} & 56.67\textsubscript{$\pm$7.97} & \textbf{83.93}\textsubscript{$\pm$5.35 \textcolor{red}{($\uparrow$27.26)}} \\
			& F1    & 88.46\textsubscript{$\pm$8.04} & \textbf{95.79}\textsubscript{$\pm$3.15 \textcolor{red}{($\uparrow$7.33)}} & 75.47\textsubscript{$\pm$12.46} & \textbf{93.16}\textsubscript{$\pm$4.46 \textcolor{red}{($\uparrow$17.69)}} & 60.22\textsubscript{$\pm$19.63} & \textbf{87.88}\textsubscript{$\pm$6.35 \textcolor{red}{($\uparrow$27.66)}} & 20.71\textsubscript{$\pm$22.54} & \textbf{80.48}\textsubscript{$\pm$8.36 \textcolor{red}{($\uparrow$59.77)}} \\
			\midrule[0.6pt]
			
			\multirow{2}*{SSLS} & Acc   & 90.07\textsubscript{$\pm$5.77} & \textbf{96.37}\textsubscript{$\pm$2.67 \textcolor{red}{($\uparrow$6.30)}} & 82.59\textsubscript{$\pm$8.48} & \textbf{93.61}\textsubscript{$\pm$3.66 \textcolor{red}{($\uparrow$11.02)}} & 72.66\textsubscript{$\pm$13.18} & \textbf{89.26}\textsubscript{$\pm$4.74 \textcolor{red}{($\uparrow$16.60)}} & 56.19\textsubscript{$\pm$5.89} & \textbf{84.22}\textsubscript{$\pm$4.58 \textcolor{red}{($\uparrow$28.03)}} \\
			& F1    & 88.55\textsubscript{$\pm$7.86} & \textbf{96.24}\textsubscript{$\pm$2.91 \textcolor{red}{($\uparrow$7.69)}} & 77.72\textsubscript{$\pm$13.38} & \textbf{93.23}\textsubscript{$\pm$4.29 \textcolor{red}{($\uparrow$15.51)}} & 57.90\textsubscript{$\pm$28.53} & \textbf{87.81}\textsubscript{$\pm$6.33 \textcolor{red}{($\uparrow$29.91)}} & 20.53\textsubscript{$\pm$16.43} & \textbf{81.46}\textsubscript{$\pm$6.10 \textcolor{red}{($\uparrow$60.93)}} \\
			\midrule[0.6pt]
			
			\multirow{2}*{LSFLS} & Acc   & 93.94\textsubscript{$\pm$2.80} & \textbf{96.43}\textsubscript{$\pm$2.38 \textcolor{red}{($\uparrow$2.49)}} & 83.58\textsubscript{$\pm$6.92} & \textbf{93.28}\textsubscript{$\pm$4.22 \textcolor{red}{($\uparrow$9.70)}} & 74.85\textsubscript{$\pm$8.56} & \textbf{89.91}\textsubscript{$\pm$5.26 \textcolor{red}{($\uparrow$15.06)}} & 59.10\textsubscript{$\pm$5.80} & \textbf{82.78}\textsubscript{$\pm$6.60 \textcolor{red}{($\uparrow$23.68)}} \\
			& F1    & 93.48\textsubscript{$\pm$3.17} & \textbf{96.30}\textsubscript{$\pm$2.67 \textcolor{red}{($\uparrow$2.82)}} & 79.55\textsubscript{$\pm$10.29} & \textbf{92.77}\textsubscript{$\pm$5.00 \textcolor{red}{($\uparrow$13.22)}} & 64.83\textsubscript{$\pm$15.54} & \textbf{88.43}\textsubscript{$\pm$6.85 \textcolor{red}{($\uparrow$23.60)}} & 29.11\textsubscript{$\pm$16.76} & \textbf{78.48}\textsubscript{$\pm$9.95 \textcolor{red}{($\uparrow$49.37)}} \\
			\midrule[0.6pt]
			
			\textbf{Avg.}& Acc&N/A&96.22\textsubscript{$\pm$2.53}&N/A&93.44\textsubscript{$\pm$3.95}&N/A&89.22\textsubscript{$\pm$4.92}&N/A&83.39\textsubscript{$\pm$5.40}\\
			(+UP4LS)& F1&N/A&96.07\textsubscript{$\pm$2.84}&N/A&92.92\textsubscript{$\pm$4.67}&N/A&87.59\textsubscript{$\pm$6.66}&N/A&79.68\textsubscript{$\pm$7.97}\\
			\midrule[0.6pt]
			
			\multicolumn{2}{c}{\textbf{Avg. ($\mathbf{\Delta}$)}}&\multicolumn{2}{c}{Acc: \textcolor{red}{$\uparrow$ 21.36} || F1: \textcolor{red}{$\uparrow$ 41.74}}& \multicolumn{2}{c}{Acc: \textcolor{red}{$\uparrow$ 23.99} || F1: \textcolor{red}{$\uparrow$ 46.32}}& \multicolumn{2}{c}{Acc: \textcolor{red}{$\uparrow$ 25.19} || F1: \textcolor{red}{$\uparrow$ 50.98}}& \multicolumn{2}{c}{Acc: \textcolor{red}{$\uparrow$ 28.99} || F1: \textcolor{red}{$\uparrow$ 65.59}}\\
			\bottomrule[1pt]
		\end{tabular}%
	}
\end{table*}%

\paragraph {Hyperparameters.} UP4LS uses the ``Bert-base-cased'' model. ${\gamma}$ is 5, the topic number of the LDA is 2. The detailed hyperparameter settings of the ``Encoder'' can be found in the corresponding papers. Adam \cite{Adam} is employed with an initial learning rate of 5e-5.

\paragraph {Evaluation metrics.} Accuracy (Acc) and the F1 score are used to evaluate the models' performance.

\begin{equation}
\begin{gathered}
{\text{Acc}} = \frac{{{\text{TP + TN}}}}{{{\text{TP + FP + TN + FN}}}}, \\ 
{\text{F1}} = {{2 \times ({\text{P}} \times {\text{R}})} \mathord{\left/
		{\vphantom {{2 \times ({\text{P}} \times {\text{R}})} {({\text{P}} + }}} \right.
		\kern-\nulldelimiterspace} {({\text{P}} + }}{\text{R}}), \\ 
\end{gathered}
\end{equation}

\noindent where, TP, FP, TN, and FN are the quantity of true positive, false positive, true negative, and false negative examples. P and R are precision and recall.


\subsection{Comparison experiments}\label{sec32}
\subsubsection{LS-task baselines}\label{sec321}

\begin{figure}[!htbp]
	\centering
	\includegraphics[width=0.44\textwidth]{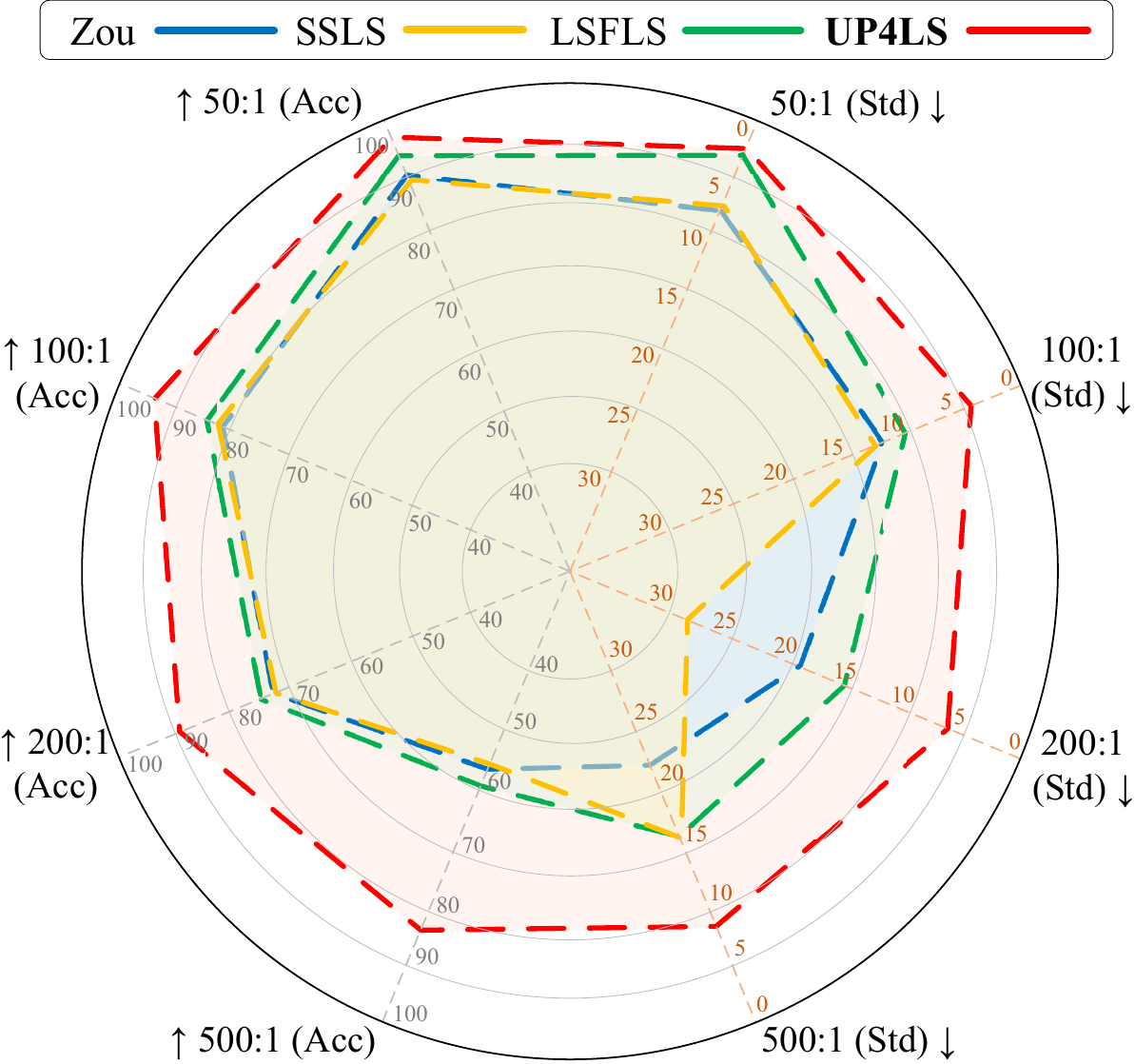}
	\caption{Comparison between the original BERT-based \textbf{LS-task} baselines and with UP4LS. For clarity, the UP4LS performance is shown here as the average of these baselines with UP4LS (``Avg.(+UP4LS)'' in Table \ref{ls-task}). The lower the Std of F1, the more stable the performance on different data. The scale on the right half is opposite to that on the left half. The larger the overall presentation area, the better the performance.}
	\label{ls-task-fig}
	\vspace{-2ex}
\end{figure}

Table \ref{ls-task} shows the comparison between the original LS-task baselines and with UP4LS. We use the Acc value and F1 Std (standard deviation) of BERT-based LS baselines in Table \ref{ls-task} to make Figure \ref{ls-task-fig}. Figure \ref{ls-task-fig} shows the performance in different aspects. Since the non-BERT-based baselines have lower Acc and F1 \cite{FEFT}\cite{TS_BiRNN}, they are not shown in Figure 6.

The results of Figure \ref{ls-task-fig} and Table \ref{ls-task} show that: 
\begin{itemize}
	\item UP4LS can improve the performance of the LS-task baselines. The Acc and F1 improvement reached 28.99\% and 65.59\% in 500:1. 
	\item The improvement increases with the increase of the ratio. In the datasets with extremely large ratios (cover:stego=500:1), the improvement is the most. The reason is that UP4LS captures user features. This shows that the advantage of UP4LS is that there are few stego, which are difficult to detect with existing methods. It can effectively capture the distributions in the few stego.
	\item UP4LS performs more stably on different user datasets. The standard deviation of the original BERT-based baselines is higher after using the UP4LS proposed.
\end{itemize}


\subsubsection{Related-task baselines}\label{sec322}
\begin{figure}[!htbp]
	\centering
	\includegraphics[width=0.47\textwidth]{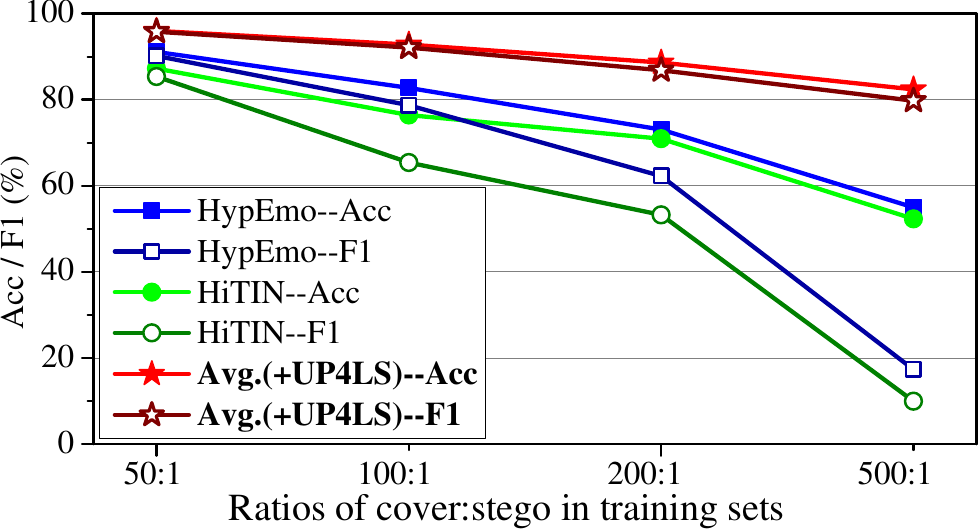}
	\caption{Comparison between the original \textbf{related-task} baselines and with UP4LS. For clarity, the UP4LS' performance is shown here as the average of these baselines with UP4LS in Table \ref{related-task}. The vertical axis is performance. The complete data are shown in Table \ref{a-5} in Appendix \ref{appendix}.}
	\label{related-task-fig}
	\vspace{-2.5ex}
\end{figure}

Table \ref{related-task} shows the comparison between the original related-task baselines and with UP4LS. We use the Acc and F1 value in Table \ref{related-task} to make Figure \ref{related-task-fig}.

\begin{table}[!h]
	\centering
	{\fontsize{7.5pt}{9pt}\selectfont
		\setlength{\tabcolsep}{0.1mm}
		\caption{Overall comparison of the original \textbf{related-task} baselines (Original) and with UP4LS (+UP4LS) in the distinct datasets. The meaning of ``a\textsubscript{$\pm$b \textcolor{red}{(c)}}'', $\Delta$, and \textbf{Bold} are the same as Table \ref{ls-task}. The Unit is \%. The complete data are shown in Table \ref{a-5} in Appendix \ref{appendix}.}\label{related-task}
		\begin{tabular}{ccccccc}
			\toprule[1pt]
			\multicolumn{2}{c}{\multirow{1.5}*{\textbf{Related-task}}} & \multicolumn{2}{c}{HypEmo \cite{HypEmo}} & \multicolumn{2}{c}{HiTIN \cite{HiTIN}} \\ 
			\multicolumn{2}{c}{\textbf{(\%)}} & Original & +UP4LS & Original & +UP4LS\\ 
			\midrule[0.6pt]
			
			\multirow{2}*{50:1} & Acc & 91.08\textsubscript{$\pm$2.32}  & \textbf{95.87}\textsubscript{$\pm$2.88 \textcolor{red}{($\uparrow$4.79)}}  &87.20\textsubscript{$\pm$5.99}&\textbf{95.97}\textsubscript{$\pm$2.32 \textcolor{red}{($\uparrow$8.77)}}\\ 
			~ & F1 & 90.15\textsubscript{$\pm$2.80}  & \textbf{95.70}\textsubscript{$\pm$3.21 \textcolor{red}{($\uparrow$5.55)}}  & 85.34\textsubscript{$\pm$7.77}&\textbf{95.82}\textsubscript{$\pm$2.46 \textcolor{red}{($\uparrow$10.48)}}  \\ 
			\midrule[0.6pt]
			
			\multirow{2}*{100:1} & Acc & 82.69\textsubscript{$\pm$5.92}  & \textbf{92.84}\textsubscript{$\pm$4.46 \textcolor{red}{($\uparrow$10.15)}} &76.40\textsubscript{$\pm$13.96}&\textbf{92.67}\textsubscript{$\pm$4.56 \textcolor{red}{($\uparrow$16.27)}}  \\
			~ & F1 &78.70\textsubscript{$\pm$8.46}  & \textbf{92.24}\textsubscript{$\pm$5.08 \textcolor{red}{($\uparrow$13.54)}}  &65.39\textsubscript{$\pm$24.96}&\textbf{91.89}\textsubscript{$\pm$5.82 \textcolor{red}{($\uparrow$26.50)}} \\ 
			\midrule[0.6pt]
			
			\multirow{2}*{200:1} & Acc & 73.05\textsubscript{$\pm$6.10}  & \textbf{88.11}\textsubscript{$\pm$4.94 \textcolor{red}{($\uparrow$15.06)}}  &70.90\textsubscript{$\pm$14.80}&\textbf{89.04}\textsubscript{$\pm$4.86 \textcolor{red}{($\uparrow$18.14)}} \\
			~ & F1 & 62.26\textsubscript{$\pm$11.40}  & \textbf{86.59}\textsubscript{$\pm$5.76 \textcolor{red}{($\uparrow$24.33)}}  &53.22\textsubscript{$\pm$31.43}&\textbf{87.00}\textsubscript{$\pm$7.06 \textcolor{red}{($\uparrow$33.78)}} \\ 
			\midrule[0.6pt]
			
			\multirow{2}*{500:1} & Acc & 54.98\textsubscript{$\pm$3.91}  & \textbf{81.84}\textsubscript{$\pm$7.36 \textcolor{red}{($\uparrow$26.86)}} &52.30\textsubscript{$\pm$1.72}&\textbf{82.91}\textsubscript{$\pm$5.21 \textcolor{red}{($\uparrow$30.61)}}  \\ 
			~ & F1 & 17.35\textsubscript{$\pm$12.16}  & \textbf{78.55}\textsubscript{$\pm$10.83 \textcolor{red}{($\uparrow$61.20)}}  &9.97\textsubscript{$\pm$9.66}&\textbf{80.89}\textsubscript{$\pm$6.17 \textcolor{red}{($\uparrow$70.92)}} \\
			\bottomrule[1pt]
		\end{tabular}
	}
\end{table}

The results of Figure \ref{related-task-fig} and Table \ref{related-task} show that: UP4LS can also help the related-task baselines perform LS in various data, and the degree of improvement increases with the increase of ratio.

All comparison experiments used ``T'' and ``Mann–Whitney U'' test, and the results are shown in Table \ref{stest} in Appendix \ref{appendix}. The results are all lower than 0.05, which is incompatible with the null hypothesis. It shows that the results are statistically significant.


\subsection{Ablation experiment}\label{sec33}
As the main contributions of this paper, we explored the effect of ``\textbf{user features}'' and ``\textbf{attention fusion}'', and we conducted this experiment.

\paragraph {User features.} we compare the performance of content features with that of ``user+content'' features. Table \ref{Ablation1} shows the comparison without and with user features. 

\begin{table}[!htbp]
	\centering
	\small
	\setlength{\tabcolsep}{2.5mm}
	\renewcommand\arraystretch{1}
	\caption{Ablation experiment of the user features. The complete data are shown in Table \ref{a-a} in Appendix \ref{appendix}.}\label{Ablation1}
	\begin{tabular}{cccc}
		\toprule[1pt]
		\multicolumn{2}{c}{\textbf{User features (\%)}} & Content&Content+User  \\ 
		\midrule[0.6pt]
		
		\multirow{2}*{50:1} & Acc & 91.34\textsubscript{$\pm$4.49}  & \textbf{96.22}\textsubscript{$\pm$2.53 \textcolor{red}{($\uparrow$ 4.88)}} \\ 
		~ & F1 & 90.17\textsubscript{$\pm$5.98}&\textbf{96.07}\textsubscript{$\pm$2.84 \textcolor{red}{($\uparrow$ 5.90)}}  \\ 
		\midrule[0.6pt]
		
		\multirow{2}*{100:1} & Acc & 82.38\textsubscript{$\pm$7.32}&\textbf{93.44}\textsubscript{$\pm$3.95 \textcolor{red}{($\uparrow$ 11.06)}}  \\
		~ & F1 &77.58\textsubscript{$\pm$11.47}& \textbf{92.92}\textsubscript{$\pm$4.67 \textcolor{red}{($\uparrow$ 15.34)}} \\ 
		\midrule[0.6pt]
		
		\multirow{2}*{200:1} & Acc & 73.38\textsubscript{$\pm$9.64} & \textbf{89.22}\textsubscript{$\pm$4.92 \textcolor{red}{($\uparrow$ 15.84)}} \\
		~ & F1 & 60.98\textsubscript{$\pm$19.77}& \textbf{87.59}\textsubscript{$\pm$6.66 \textcolor{red}{($\uparrow$ 26.61)}} \\ 
		\midrule[0.6pt]
		
		\multirow{2}*{500:1} & Acc & 57.32\textsubscript{$\pm$5.54}&\textbf{83.39}\textsubscript{$\pm$5.40 \textcolor{red}{($\uparrow$ 26.07)}}  \\ 
		~ & F1 & 23.45\textsubscript{$\pm$15.74} &  \textbf{79.68}\textsubscript{$\pm$7.97 \textcolor{red}{($\uparrow$ 56.23)}} \\ 
		\midrule[0.6pt]
		
		\textbf{Avg. ($\mathbf{\Delta}$)} & \multicolumn{3}{c}{Acc: \textcolor{red}{$\uparrow$ 14.46} \quad || \quad F1: \textcolor{red}{$\uparrow$ 26.02}}  \\
		\bottomrule[1pt]
	\end{tabular}
	\vspace{-2.5ex}
\end{table}

The results of Table \ref{Ablation1} show that: As the ratio increases, the degree of improvement shows an increasing trend. This is attributed to user features reflecting the user's style to a certain extent. Even with a few quantity of stegos, more comprehensive user features can be captured. Therefore, the user feature has a stable performance.

\paragraph {Attention fusion.} We compare the impact of using mutual attention mechanism to guide feature fusion and simple concatenating on detection performance, as shown in Table 6.

\begin{table}[htbp]
	\centering
	\small
	\renewcommand\arraystretch{1}
	\caption{Ablation experiment of the attention fusion. Take the ratio of 50:1 as an example. The complete data are shown in Table \ref{a-b} in Appendix \ref{appendix}.}\label{Ablation2}%
	\begin{tabular}{ccc}
		\toprule[1pt]
		\textbf{Attention fusion (\%)}& Concat&Attn \\
		\midrule[0.6pt]
		Acc   & 95.88\textsubscript{$\pm$2.09} &\textbf{96.76}\textsubscript{$\pm$2.00 \textcolor{red}{($\uparrow$ 0.88)}} \\
		F1    &95.56\textsubscript{$\pm$2.55} &\textbf{96.67}\textsubscript{$\pm$2.11 \textcolor{red}{($\uparrow$ 1.11)}} \\
		\bottomrule[1pt]
	\end{tabular}%
\end{table}%

From the results in Table \ref{Ablation2}, it can be seen that the attention can further enhance the feature expression and improve the detection performance.

\section{Conclusion}\label{sec5}
In this paper, we propose UP4LS, which constructs the user profile for enhancing LS. UP4LS has explored three types of user attributes and extracted user features by the designed extraction modules. Existing methods retain the designed deep-learning model and add UP4LS to other parts to improve their performance in complex realistic scenarios. Experiments show that UP4LS can significantly enhance the performance of LS-task and related-task SOTA baselines in social networks. Especially when there are very few stego samples. And the detection stability in various data is enhanced.

In the future, we will design LS with user behavior. It detects covert communications more directly. In addition, stegos in social networks may be generated and mixed by multiple steganography. There is little research on the detection of these stegos. Therefore, we also research the steganography algorithm rather than the stego as the detection object.

\newpage
\section*{Limitations}
This paper constructs the user profile and extracts user features that are beneficial to detect stegos. While this research improves the performance of existing methods, it still faces certain limitations and potential risks: 

\textbf{(1) Use of language model:} The language model of the text is not designed too much or uses LLMs such as LLaMA3. This is because the design focus of this paper is the construction of user profile and the extraction of user features. If a larger pre-trained model is used to extract content features, it may indeed further improve the detection capability. 

\textbf{(2) User profile completeness:} Although we strive to comprehensively analyze user attributes, the given user profile may not encompass all aspects like user metadata. Moreover, exploring extraction from other user behaviors could potentially uncover additional attributes beneficial to LS. 

\textbf{(3) The broad advantage in ideal data:} In ideal data, UP4LS has potential risks in improving performance. There are slight or even no user attributes reflected in these data. User features hardly improve the performance of these data.

\section*{Ethical Statement}
This study involves collecting and analyzing publicly visible Twitter user tweets to build user portraits, aiming to study whether the text contains secret information. In this study, we promise: 

\textbf{(1) Data Collection:} All collected data comes from the publicly accessible Twitter platform and contains only non-sensitive information. We will not collect any information that can directly identify individuals. 

\textbf{(2) Data Use:} The collected data will only be used for scientific research purposes, that is, to detect whether the text is steganographic. It will not be used for any commercial purpose. 

\textbf{{3} Data Protection:} All data during the research process are stored in an encrypted and protected server and can only be accessed by authorized researchers. 

\textbf{{4} Research results:} In any research results released to the public, we will not disclose any information about specific users, and ensure that the presentation of research results will not cause any harm or inconvenience to any user.

\section*{Acknowledgements}\label{sec6}
This work is supported by the National Natural Science Foundation of China (Grant U21B2020) and supported by BUPT Excellent Ph.D. Students Foundation (Grant CX2023120).

\bibliography{references}\label{sec7}
\bibliographystyle{acl_natbib}

\appendix


\section{Appendix}
\label{appendix}

\begin{table*}[!htbp]
	\centering
	{\fontsize{8.5pt}{10.5pt}\selectfont
		\setlength{\tabcolsep}{1.5mm}
		\caption{The performance of the \textbf{LS-task} baselines and with UP4LS in \textbf{50:1} ratio. The meaning of ``a\textsubscript{$\pm$b \textcolor{red}{(c)}}'', $\Delta$, and \textbf{Bold} in the Table \ref{a-1} to Table \ref{a-b} are the same as Table \ref{ls-task}.}\label{a-1}
		\begin{tabular}{cccccccccccccc}
			\toprule[1.3pt]
			\multicolumn{3}{c}{\textbf{50:1 (\%)}} & U1 & U2 & U3 & U4 & U5 & U6 & U7 & U8 & U9 & U10 & Avg\textsubscript{$\pm$Std} \textsubscript{\textcolor{red}{\textbf{($\mathbf{\Delta}$Acc)}}}\\ \midrule[0.7pt]
			
			\multirow{4}*{FETS} & \multirow{2}*{Acc} & Original & 50.09  & 50.00  & 50.27  & 50.52  & 50.47  & 50.00  & 50.24  & 50.26  & 50.00  & 50.00  & 50.19\textsubscript{$\pm$0.20}  \\ 
			~ &  & +UP4LS & 95.09  & 95.98  & 96.26  & 96.91  & 96.24  & 98.12  & 87.20  & 95.06  & 97.28  & 97.77  & \textbf{95.59\textsubscript{$\pm$3.12}} \textsubscript{\textcolor{red}{\textbf{($\uparrow$ 45.40)}}}  \\ 
			\cdashline{2-14}[3pt/2pt]
			~ & \multirow{2}*{F1} & Original & 0.34  & 0.00  & 1.09  & 2.04  & 1.86  & 0.00  & 0.96  & 1.02  & 0.00  & 0.00  & 0.73\textsubscript{$\pm$0.78}  \\
			~ & ~ & +UP4LS & 95.12  & 95.86  & 96.14  & 96.96  & 96.09  & 98.14  & 85.40  & 94.81  & 97.29  & 97.72  & \textbf{95.35\textsubscript{$\pm$3.66}} \textsubscript{\textcolor{red}{\textbf{($\uparrow$ 94.62)}}}  \\ \midrule[0.7pt]
			
			\multirow{4}*{TS\_RNN} & \multirow{2}*{Acc} & Original & 50.00 & 50.52 & 50.00 & 50.10 & 50.12 & 50.08 & 50.00 & 50.09 & 50.11 & 50.09 & 50.11\textsubscript{$\pm$0.15} \\ 
			~ & & +UP4LS & 96.90 & 96.50 & 97.72 & 96.70 & 97.76 & 98.86 & 91.67 & 95.66 & 97.93 & 97.86 & \textbf{96.76\textsubscript{$\pm$2.00}} \textsubscript{\textcolor{red}{\textbf{($\uparrow$ 46.65)}}} \\ 
			\cdashline{2-14}[3pt/2pt]
			~ & \multirow{2}*{F1} & Original & 0.00 & 2.08 & 0.00 & 0.41 & 0.47 & 0.33 & 0.00 & 0.34 & 0.43 & 0.36 & 0.44\textsubscript{$\pm$0.61} \\
			~ & ~ & +UP4LS & 96.90 & 96.43 & 97.68 & 96.60 & 97.71 & 98.85 & 91.28 & 95.51 & 97.90 & 97.81 & \textbf{96.67\textsubscript{$\pm$2.11}} \textsubscript{\textcolor{red}{\textbf{($\uparrow$ 96.23)}}} \\ \midrule[0.7pt]
			
			\multirow{4}*{Zou} & \multirow{2}*{Acc} & Original & 88.02  & 93.36  & 85.40  & 92.06  & 91.41  & 93.88  & 75.72  & 92.08  & 92.37  & 95.80  & 90.01\textsubscript{$\pm$5.82}  \\ 
			~ &  & +UP4LS & 95.74  & 97.68  & 95.44  & 97.01  & 97.06  & 98.22  & 88.41  & 95.83  & 97.17  & 96.96  & \textbf{95.95\textsubscript{$\pm$2.79}} \textsubscript{\textcolor{red}{\textbf{($\uparrow$ 5.94)}}} \\ 
			\cdashline{2-14}[3pt/2pt]
			~ & \multirow{2}*{F1} & Original & 86.47  & 92.90  & 82.91  & 91.38  & 90.63  & 93.50  & 68.04  & 91.44  & 91.75  & 95.62  & 88.46\textsubscript{$\pm$8.04}  \\ 
			~ & ~ & +UP4LS & 95.76  & 97.67  & 95.25  & 97.03  & 97.03  & 98.22  & 87.20  & 95.70  & 97.09  & 96.90  & \textbf{95.79\textsubscript{$\pm$3.15}} \textsubscript{\textcolor{red}{\textbf{($\uparrow$ 7.33)}}} \\ \midrule[0.7pt]
			
			\multirow{4}*{SSLS} & \multirow{2}*{Acc} & Original & 90.09  & 94.93  & 93.25  & 87.22  & 88.94  & 95.60  & 76.21  & 88.50  & 90.20  & 95.71  & 90.07\textsubscript{$\pm$5.77}  \\ 
			~ &  & +UP4LS & 95.67  & 97.53  & 95.75  & 97.34  & 97.76  & 98.24  & 89.20  & 97.00  & 97.10  & 98.12  & \textbf{96.37\textsubscript{$\pm$2.67}} \textsubscript{\textcolor{red}{\textbf{($\uparrow$ 6.30)}}} \\ 
			\cdashline{2-14}[3pt/2pt]
			~ & \multirow{2}*{F1}  & Original & 89.20  & 94.68  & 92.76  & 85.34  & 87.63  & 95.39  & 68.78  & 87.08  & 89.13  & 95.53  & 88.55\textsubscript{$\pm$7.86}  \\ 
			~ & ~ & +UP4LS & 95.55  & 97.50  & 95.58  & 97.30  & 97.73  & 98.25  & 88.38  & 96.99  & 97.01  & 98.12  & \textbf{96.24\textsubscript{$\pm$2.91}} \textsubscript{\textcolor{red}{\textbf{($\uparrow$ 7.69)}}} \\ \midrule[0.7pt]
			
			\multirow{4}*{LSFLS} & \multirow{2}*{Acc} & Original & 90.28  & 95.72  & 93.12  & 92.27  & 94.35  & 97.88  & 88.89  & 95.40  & 95.10  & 96.34  & 93.94\textsubscript{$\pm$2.80}  \\
			~ &  & +UP4LS & 94.40  & 97.52  & 97.45  & 97.22  & 96.47  & 98.09  & 90.34  & 97.31  & 97.80  & 97.70  & \textbf{96.43\textsubscript{$\pm$2.38}} \textsubscript{\textcolor{red}{\textbf{($\uparrow$ 2.49)}}} \\  
			\cdashline{2-14}[3pt/2pt]
			~ & \multirow{2}*{F1} & Original & 89.18  & 95.53  & 92.56  & 91.68  & 94.01  & 97.84  & 87.80  & 95.18  & 94.85  & 96.21  & 93.48\textsubscript{$\pm$3.17} \\ 
			~ & ~ & +UP4LS & 94.22  & 97.48  & 97.45  & 97.17  & 96.51  & 98.08  & 89.36  & 97.30  & 97.76  & 97.68  & \textbf{96.30\textsubscript{$\pm$2.67}} \textsubscript{\textcolor{red}{\textbf{($\uparrow$ 2.82)}}} \\ \bottomrule[1.3pt]	
		\end{tabular}
	}
\end{table*}

\begin{table*}[!ht]
	\centering
	{\fontsize{8.5pt}{10.5pt}\selectfont
		\setlength{\tabcolsep}{1.5mm}
		\caption{The performance of the \textbf{LS-task} baselines and with UP4LS in \textbf{100:1} ratio.}\label{a-2}
		\begin{tabular}{cccccccccccccc}
			\toprule[1.3pt]
			\multicolumn{3}{c}{\textbf{100:1 (\%)}} & U1 & U2 & U3 & U4 & U5 & U6 & U7 & U8 & U9 & U10 & Avg\textsubscript{$\pm$Std} \textsubscript{\textcolor{red}{\textbf{($\mathbf{\Delta}$Acc)}}} \\ 
			\midrule[0.7pt]
			
			\multirow{4}*{FETS} & \multirow{2}*{Acc} & Original & 50.00  & 50.00  & 50.00  & 50.00  & 50.00  & 50.00  & 50.12  & 50.00  & 50.00  & 50.00  & 50.01\textsubscript{$\pm$0.04}  \\ 
			~ &  & +UP4LS & 94.19  & 95.80  & 94.01  & 94.62  & 93.79  & 95.02  & 81.16  & 93.10  & 94.34  & 94.77  & \textbf{93.08\textsubscript{$\pm$4.25}} \textsubscript{\textcolor{red}{\textbf{($\uparrow$ 43.07)}}} \\ 
			\cdashline{2-14}[3pt/2pt]
			~ & \multirow{2}*{F1} & Original & 0.00  & 0.00  & 0.00  & 0.00  & 0.00  & 0.00  & 0.48  & 0.00  & 0.00  & 0.00  & 0.05\textsubscript{$\pm$0.15}  \\ 
			~ & ~ & +UP4LS & 94.19  & 95.62  & 93.56  & 94.29  & 93.36  & 94.77  & 77.90  & 93.40  & 94.00  & 94.49  & \textbf{92.56\textsubscript{$\pm$5.20}}  \textsubscript{\textcolor{red}{\textbf{($\uparrow$ 92.51)}}}\\ \midrule[0.7pt]
			
			\multirow{4}*{TS\_RNN} & \multirow{2}*{Acc} & Original & 50.00 & 50.17 & 50.00 & 50.10 & 50.00 & 50.00 & 50.12 & 50.00 & 50.00 & 50.09 & 50.05\textsubscript{$\pm$0.07} \\ 
			~ & & +UP4LS & 93.10 & 96.24 & 94.18 & 95.88 & 94.71 & 95.32 & 82.00 & 93.78 & 93.46 & 94.29 & \textbf{93.30\textsubscript{$\pm$4.10}} \textsubscript{\textcolor{red}{\textbf{($\uparrow$ 43.25)}}}\\ 
			\cdashline{2-14}[3pt/2pt]
			~ & \multirow{2}*{F1}  & Original & 0.00 & 0.70 & 0.00 & 0.41 & 0.00 & 0.00 & 0.48 & 0.00 & 0.00 & 0.36 & 0.20\textsubscript{$\pm$0.27} \\ 
			~ & ~ & +UP4LS & 92.95 & 96.20 & 93.88 & 95.71 & 94.41 & 95.09 & 80.11 & 93.42 & 93.01 & 93.98 & \textbf{92.88\textsubscript{$\pm$4.62}} \textsubscript{\textcolor{red}{\textbf{($\uparrow$ 92.68)}}}\\ \midrule[0.7pt]
			
			\multirow{4}*{Zou} & \multirow{2}*{Acc} & Original & 75.43  & 91.70  & 80.02  & 78.76  & 74.35  & 90.78  & 66.43  & 83.05  & 85.62  & 83.75  & 80.99\textsubscript{$\pm$7.72}  \\ 
			~ &  & +UP4LS & 93.10  & 96.50  & 94.56  & 95.77  & 94.82  & 94.94  & 83.70  & 95.40  & 94.46  & 95.87  & \textbf{93.91\textsubscript{$\pm$3.71}} \textsubscript{\textcolor{red}{\textbf{($\uparrow$ 12.92)}}} \\ 
			\cdashline{2-14}[3pt/2pt]
			~ & \multirow{2}*{F1} & Original & 67.43  & 90.94  & 75.03  & 73.04  & 65.51  & 89.87  & 49.45  & 79.59  & 83.21  & 80.60  & 75.47\textsubscript{$\pm$12.46}  \\ 
			~ & ~ & +UP4LS & 92.69  & 95.39  & 93.34  & 95.86  & 93.58  & 94.69  & 80.84  & 95.34  & 94.16  & 95.68  & \textbf{93.16\textsubscript{$\pm$4.46}} \textsubscript{\textcolor{red}{\textbf{($\uparrow$ 17.69)}}} \\ \midrule[0.7pt]
			
			\multirow{4}*{SSLS} & \multirow{2}*{Acc} & Original & 79.22  & 91.35  & 80.66  & 76.70  & 87.29  & 93.56  & 66.30  & 75.13  & 86.71  & 88.93  & 82.59\textsubscript{$\pm$8.48}  \\ 
			~ &  & +UP4LS & 92.90  & 95.37  & 93.70  & 95.05  & 94.59  & 95.11  & 83.48  & 95.03  & 95.53  & 95.36  & \textbf{93.61\textsubscript{$\pm$3.66}} \textsubscript{\textcolor{red}{\textbf{($\uparrow$ 11.02)}}} \\ 
			\cdashline{2-14}[3pt/2pt]
			~ & \multirow{2}*{F1} & Original & 73.89  & 90.54  & 76.02  & 69.62  & 85.48  & 93.15  & 49.36  & 66.89  & 84.67  & 87.55  & 77.72\textsubscript{$\pm$13.38} \\ 
			~ & ~ & +UP4LS & 92.75  & 95.15  & 93.88  & 94.81  & 94.28  & 94.86  & 81.22  & 94.80  & 95.34  & 95.16  & \textbf{93.23\textsubscript{$\pm$4.29}} \textsubscript{\textcolor{red}{\textbf{($\uparrow$ 15.51)}}} \\ \midrule[0.7pt]
			
			\multirow{4}*{LSFLS} & \multirow{2}*{Acc} & Original & 79.86  & 91.78  & 80.93  & 82.89  & 77.65  & 91.68  & 70.77  & 81.09  & 89.83  & 89.29  & 83.58\textsubscript{$\pm$6.92}  \\ 
			~ & & +UP4LS & 92.67  & 95.72  & 93.47  & 94.33  & 93.51  & 94.60  & 81.67  & 95.40  & 95.58  & 95.86  & \textbf{93.28\textsubscript{$\pm$4.22}} \textsubscript{\textcolor{red}{\textbf{($\uparrow$ 9.70)}}} \\ 
			\cdashline{2-14}[3pt/2pt]
			~ &  \multirow{2}*{F1} & Original & 74.75  & 91.05  & 76.44  & 79.35  & 71.30  & 90.93  & 58.70  & 76.68  & 88.26  & 88.00  & 79.55\textsubscript{$\pm$10.29}  \\ 
			~ & ~ & +UP4LS & 92.41  & 95.59  & 93.19  & 93.98  & 93.02  & 94.28  & 78.92  & 95.22  & 95.45  & 95.63  & \textbf{92.77\textsubscript{$\pm$5.00}} \textsubscript{\textcolor{red}{\textbf{($\uparrow$ 13.22)}}} \\  \bottomrule[1.3pt]	
		\end{tabular}
	}
\end{table*}
\vspace{-3ex}

\begin{table*}[!ht]
	\centering
	{\fontsize{8.5pt}{10.5pt}\selectfont
		\setlength{\tabcolsep}{1.5mm}
		\caption{The performance of the \textbf{LS-task} baselines and with UP4LS in \textbf{200:1} ratio.}\label{a-3}
		\begin{tabular}{cccccccccccccc}
			\toprule[1.3pt]
			\multicolumn{3}{c}{\textbf{200:1 (\%)}} & U1 & U2 & U3 & U4 & U5 & U6 & U7 & U8 & U9 & U10 & Avg\textsubscript{$\pm$Std} \textsubscript{\textcolor{red}{\textbf{($\mathbf{\Delta}$Acc)}}} \\ \midrule[0.7pt]
			
			\multirow{4}*{FETS} & \multirow{2}*{Acc} & Original & 50.00  & 50.00  & 50.00  & 50.00  & 50.00  & 50.00  & 50.12  & 50.00  & 50.00  & 50.00  & 50.01\textsubscript{$\pm$0.04}  \\
			~ &  & +UP4LS & 83.88  & 92.22  & 92.96  & 88.87  & 91.29  & 91.19  & 75.02  & 92.25  & 88.89  & 91.52  & \textbf{88.81\textsubscript{$\pm$5.52}} \textsubscript{\textcolor{red}{\textbf{($\uparrow$ 38.80)}}} \\ 
			\cdashline{2-14}[3pt/2pt]
			~ & \multirow{2}*{F1} & Original & 0.00  & 0.00  & 0.00  & 0.00  & 0.00  & 0.00  & 0.48  & 0.00  & 0.00  & 0.00  & 0.05\textsubscript{$\pm$0.15}  \\ 
			~ & ~ & +UP4LS & 81.58  & 92.21  & 92.49  & 87.59  & 90.80  & 90.36  & 64.69  & 91.66  & 87.50  & 90.73  & \textbf{86.96\textsubscript{$\pm$8.47}} \textsubscript{\textcolor{red}{\textbf{($\uparrow$ 86.91)}}} \\ \midrule[0.7pt]
			
			\multirow{4}*{TS\_RNN} & \multirow{2}*{Acc} & Original & 50.00 & 50.09 & 50.00 & 50.00 & 50.00 & 50.00 & 50.00 & 50.12 & 50.00 & 50.00 & 50.02\textsubscript{$\pm$0.04} \\ 
			~ & & +UP4LS & 85.60 & 94.32 & 91.51 & 88.64 & 88.73 & 91.27 & 77.29 & 90.41 & 87.69 & 91.16 & \textbf{88.66\textsubscript{$\pm$4.67}} \textsubscript{\textcolor{red}{\textbf{($\uparrow$ 38.64)}}}\\
			\cdashline{2-14}[3pt/2pt]
			~ &  \multirow{2}*{F1} & Original & 0.00 & 0.35 & 0.00 & 0.00 & 0.00 & 0.00 & 0.00 & 0.48 & 0.00 & 0.00 & 0.08\textsubscript{$\pm$0.18} \\ 
			~ & ~ & +UP4LS & 84.23 & 94.06 & 90.75 & 85.99 & 86.85 & 90.69 & 71.25 & 88.79 & 85.96 & 90.30 & \textbf{86.89\textsubscript{$\pm$6.23}} \textsubscript{\textcolor{red}{\textbf{($\uparrow$ 86.81)}}}\\ \midrule[0.7pt]		
			
			\multirow{4}*{Zou} & \multirow{2}*{Acc} & Original & 52.76  & 75.61  & 67.34  & 72.68  & 71.29  & 83.44  & 69.79  & 73.68  & 78.00  & 81.52  & 72.61\textsubscript{$\pm$8.61}  \\
			~ &  & +UP4LS & 85.26  & 94.93  & 92.24  & 86.62  & 91.27  & 93.36  & 78.14  & 90.56  & 89.54  & 92.50  & \textbf{89.44\textsubscript{$\pm$4.95}} \textsubscript{\textcolor{red}{\textbf{($\uparrow$ 16.83)}}} \\  
			\cdashline{2-14}[3pt/2pt]
			~ & \multirow{2}*{F1} & Original & 10.46  & 67.75  & 51.49  & 62.41  & 59.74  & 80.20  & 56.72  & 64.28  & 71.79  & 77.33  & 60.22\textsubscript{$\pm$19.63}  \\ 
			~ & ~ & +UP4LS & 82.88  & 94.81  & 91.61  & 84.28  & 90.39  & 92.49  & 73.11  & 89.03  & 88.32  & 91.89  & \textbf{87.88\textsubscript{$\pm$6.35}} \textsubscript{\textcolor{red}{\textbf{($\uparrow$ 27.66)}}} \\ \midrule[0.7pt]
			
			\multirow{4}*{SSLS} & \multirow{2}*{Acc} & Original & 55.69  & 79.11  & 64.60  & 73.09  & 71.41  & 88.25  & 50.48  & 69.59  & 87.36  & 87.05  & 72.66\textsubscript{$\pm$13.18}  \\ 
			~ &  & +UP4LS & 86.03  & 95.02  & 92.06  & 88.37  & 91.60  & 92.17  & 77.85  & 88.47  & 89.43  & 91.61  & \textbf{89.26\textsubscript{$\pm$4.74}} \textsubscript{\textcolor{red}{\textbf{($\uparrow$ 16.60)}}} \\
			\cdashline{2-14}[3pt/2pt]
			~ & \multirow{2}*{F1} & Original & 20.68  & 74.38  & 45.20  & 63.19  & 59.97  & 86.72  & 1.91  & 56.30  & 85.54  & 85.13  & 57.90\textsubscript{$\pm$28.53}  \\ 
			~ & ~ & +UP4LS & 85.69  & 94.92  & 91.38  & 86.54  & 90.76  & 91.50  & 71.74  & 86.49  & 88.19  & 90.84  & \textbf{87.81\textsubscript{$\pm$6.33}} \textsubscript{\textcolor{red}{\textbf{($\uparrow$ 29.91)}}} \\ \midrule[0.7pt]
			
			\multirow{4}*{LSFLS} & \multirow{2}*{Acc} & Original & 65.78  & 82.69  & 70.44  & 72.27  & 70.47  & 83.03  & 61.84  & 71.29  & 85.73  & 85.00  & 74.85\textsubscript{$\pm$8.56}  \\ 
			~ &  & +UP4LS & 82.30  & 95.10  & 93.70  & 88.47  & 92.19  & 90.98  & 79.03  & 93.07  & 90.85  & 93.38  & \textbf{89.91\textsubscript{$\pm$5.26}} \textsubscript{\textcolor{red}{\textbf{($\uparrow$ 15.06)}}} \\ 
			\cdashline{2-14}[3pt/2pt]
			~ & \multirow{2}*{F1} & Original & 47.97  & 79.07  & 58.03  & 61.63  & 58.10  & 79.57  & 38.52  & 59.74  & 83.35  & 82.35  & 64.83\textsubscript{$\pm$15.54}  \\ 
			~ & ~ & +UP4LS & 78.22  & 95.01  & 93.92  & 86.80  & 91.53  & 89.00  & 74.52  & 92.56  & 89.93  & 92.83  & \textbf{88.43\textsubscript{$\pm$6.85}} \textsubscript{\textcolor{red}{\textbf{($\uparrow$ 23.60)}}} \\ \bottomrule[1.3pt]	
		\end{tabular}
	}
\end{table*}

\begin{table*}[!ht]
	\centering
	{\fontsize{8.5pt}{10.5pt}\selectfont
		\setlength{\tabcolsep}{1.5mm}
		\caption{The performance of the \textbf{LS-task} baselines and with UP4LS in \textbf{500:1} ratio.}\label{a-4}
		\begin{tabular}{cccccccccccccc}
			\toprule[1.3pt]
			\multicolumn{3}{c}{\textbf{500:1 (\%)}} & U1 & U2 & U3 & U4 & U5 & U6 & U7 & U8 & U9 & U10 & Avg\textsubscript{$\pm$Std} \textsubscript{\textcolor{red}{\textbf{($\mathbf{\Delta}$Acc)}}} \\ \midrule[0.7pt]
			
			\multirow{4}*{FETS} & \multirow{2}*{Acc} & Original & 50.00  & 50.00  & 50.00  & 50.00  & 50.00  & 50.00  & 50.12  & 50.00  & 50.00  & 50.00  & 50.01\textsubscript{$\pm$0.04}  \\ 
			~ &  & +UP4LS & 86.72  & 86.10  & 84.05  & 88.76  & 82.47  & 80.85  & 66.43  & 89.85  & 74.62  & 87.68  & \textbf{82.75\textsubscript{$\pm$7.27}} \textsubscript{\textcolor{red}{\textbf{($\uparrow$ 32.74)}}} \\ 
			\cdashline{2-14}[3pt/2pt]
			~ & \multirow{2}*{F1} & Original & 0.00  & 0.00  & 0.00  & 0.00  & 0.00  & 0.00  & 0.48  & 0.00  & 0.00  & 0.00  & 0.05\textsubscript{$\pm$0.15}  \\ 
			~ & ~ & +UP4LS & 83.42  & 84.24  & 81.08  & 87.74  & 78.74  & 74.95  & 52.56  & 88.94  & 65.99  & 85.98  & \textbf{78.36\textsubscript{$\pm$11.33}} \textsubscript{\textcolor{red}{\textbf{($\uparrow$ 78.31)}}} \\ \midrule[0.7pt]
			
			\multirow{4}*{TS\_RNN} & \multirow{2}*{Acc} & Original & 50.00 & 50.00 & 50.00 & 50.00 & 50.00 & 50.00 & 50.00 & 50.12 & 50.00 & 50.00 & 50.01\textsubscript{$\pm$0.04} \\ 
			~ &  & +UP4LS & 80.26 & 90.73 & 84.95 & 90.10 & 81.53 & 82.22 & 74.40 & 89.37 & 74.14 & 85.00 & \textbf{83.27\textsubscript{$\pm$5.97}} \textsubscript{\textcolor{red}{\textbf{($\uparrow$ 33.26)}}} \\ 
			\cdashline{2-14}[3pt/2pt]
			~ & \multirow{2}*{F1} & Original & 0.00 & 0.00 & 0.00 & 0.00 & 0.00 & 0.00 & 0.00 & 0.48 & 0.00 & 0.00 & 0.05\textsubscript{$\pm$0.15} \\ 
			~ & ~ & +UP4LS & 76.70 & 89.87 & 82.72 & 90.98 & 77.34 & 78.42 & 70.80 & 87.97 & 58.82 & 82.39 & \textbf{79.60\textsubscript{$\pm$9.70}} \textsubscript{\textcolor{red}{\textbf{($\uparrow$ 79.55)}}}\\ \midrule[0.7pt]
			
			\multirow{4}*{Zou} & \multirow{2}*{Acc} & Original & 50.69  & 66.35  & 51.82  & 50.52  & 54.00  & 72.35  & 51.09  & 54.94  & 50.65  & 64.29  & 56.67\textsubscript{$\pm$7.97}  \\
			~ &  & +UP4LS & 82.50  & 88.72  & 83.03  & 84.35  & 81.91  & 80.49  & 72.28  & 88.76  & 86.06  & 91.16  & \textbf{83.93\textsubscript{$\pm$5.35}} \textsubscript{\textcolor{red}{\textbf{($\uparrow$ 27.26)}}} \\ 
			\cdashline{2-14}[3pt/2pt]
			~ & \multirow{2}*{F1} & Original & 2.72  & 49.41  & 7.04  & 2.04  & 14.81  & 61.78  & 4.26  & 17.98  & 2.58  & 44.44  & 20.71\textsubscript{$\pm$22.54}  \\ 
			~ & ~ & +UP4LS & 79.31  & 87.44  & 80.38  & 81.45  & 77.60  & 75.40  & 61.02  & 87.78  & 83.92  & 90.53  & \textbf{80.48\textsubscript{$\pm$8.36}} \textsubscript{\textcolor{red}{\textbf{($\uparrow$ 59.77)}}} \\ \midrule[0.7pt]
			
			\multirow{4}*{SSLS} & \multirow{2}*{Acc} & Original & 50.86  & 56.91  & 52.37  & 55.26  & 54.71  & 70.64  & 53.26  & 51.62  & 55.01  & 61.25  & 56.19\textsubscript{$\pm$5.89}  \\ 
			~ &  & +UP4LS & 81.72  & 89.51  & 79.84  & 89.69  & 82.24  & 82.14  & 78.62  & 88.33  & 80.17  & 89.91  & \textbf{84.22\textsubscript{$\pm$4.58}} \textsubscript{\textcolor{red}{\textbf{($\uparrow$ 28.03)}}} \\ 
			\cdashline{2-14}[3pt/2pt]
			~ & \multirow{2}*{F1} & Original & 3.39  & 24.27  & 9.06  & 19.03  & 17.20  & 58.53  & 12.64  & 6.27  & 18.22  & 36.73  & 20.53\textsubscript{$\pm$16.43}  \\ 
			~ & ~ & +UP4LS & 79.73  & 88.39  & 74.80  & 88.91  & 78.52  & 78.51  & 74.75  & 86.91  & 75.27  & 88.80  & \textbf{81.46\textsubscript{$\pm$6.10}} \textsubscript{\textcolor{red}{\textbf{($\uparrow$ 60.93)}}} \\  \midrule[0.7pt]
			
			\multirow{4}*{LSFLS} & \multirow{2}*{Acc} & Original & 55.00  & 63.55  & 59.07  & 53.51  & 65.29  & 64.19  & 50.36  & 67.72  & 55.12  & 57.14  & 59.10\textsubscript{$\pm$5.80}  \\ 
			~ &  & +UP4LS & 81.21  & 88.46  & 84.76  & 86.08  & 80.59  & 76.75  & 68.12  & 87.56  & 83.51  & 90.71  & \textbf{82.78\textsubscript{$\pm$6.60}} \textsubscript{\textcolor{red}{\textbf{($\uparrow$ 23.68)}}} \\ 
			\cdashline{2-14}[3pt/2pt]
			~ & \multirow{2}*{F1} & Original & 19.69  & 42.64  & 27.99  & 12.40  & 46.85  & 44.22  & 1.44  & 52.33  & 18.58  & 25.00  & 29.11\textsubscript{$\pm$16.76}  \\ 
			~ & ~ & +UP4LS & 77.62  & 82.60  & 82.02  & 84.39  & 75.91  & 69.71  & 55.41  & 85.80  & 80.46  & 90.85  & \textbf{78.48\textsubscript{$\pm$9.95}} \textsubscript{\textcolor{red}{\textbf{($\uparrow$ 49.37)}}} \\  \bottomrule[1.3pt]
		\end{tabular}
	}
\end{table*}

\begin{table*}[!ht]
	\centering
	{\fontsize{8.5pt}{9.5pt}\selectfont
		\setlength{\tabcolsep}{1.5mm}
		\caption{The performance of the original \textbf{related-task} baselines and with UP4LS.}\label{a-5}
		\begin{tabular}{cccccccccccccc}
			\toprule[1.3pt]
			\multicolumn{3}{c}{HypEmo \cite{HypEmo}} & U1 & U2 & U3 & U4 & U5 & U6 & U7 & U8 & U9 & U10 & Avg\textsubscript{$\pm$Std} \textsubscript{\textcolor{red}{\textbf{($\mathbf{\Delta}$Acc)}}} \\ \midrule[0.7pt]
			\multirow{4}*{50:1} & \multirow{2}*{Acc} & Original & 88.53  & 94.93  & 92.70  & 91.44  & 89.76  & 93.23  & 86.96  & 91.74  & 90.41  & 91.07  & 91.08\textsubscript{$\pm$2.32}  \\
			~ &  & +UP4LS & 94.61  & 96.90  & 95.67  & 96.34  & 97.41  & 97.96  & 88.16  & 96.93  & 97.17  & 97.54  & \textbf{95.87\textsubscript{$\pm$2.88}} \textsubscript{\textcolor{red}{\textbf{($\uparrow$ 4.79)}}} \\
			\cdashline{2-14}[3pt/2pt]
			~ & \multirow{2}*{F1} & Original & 87.05  & 94.67  & 92.13  & 90.64  & 88.60  & 92.74  & 85.04  & 90.99  & 89.40  & 90.20  & 90.15\textsubscript{$\pm$2.80}  \\ 
			~ & ~ & +UP4LS & 94.52  & 96.84  & 95.53  & 96.22  & 97.39  & 97.92  & 87.04  & 96.91  & 97.10  & 97.51  & \textbf{95.70\textsubscript{$\pm$3.21}} \textsubscript{\textcolor{red}{\textbf{($\uparrow$ 5.55)}}} \\  \midrule[0.7pt]
			
			\multirow{4}*{100:1} & \multirow{2}*{Acc} & Original & 81.55  & 91.70  & 79.01  & 75.26  & 79.06  & 90.86  & 75.48  & 82.37  & 88.34  & 83.30  & 82.69\textsubscript{$\pm$5.92}  \\ 
			~ &  & +UP4LS & 92.23  & 95.54  & 91.98  & 93.99  & 94.44  & 94.15  & 80.80  & 93.53  & 96.51  & 95.25  & \textbf{92.84\textsubscript{$\pm$4.46}} \textsubscript{\textcolor{red}{\textbf{($\uparrow$ 10.15)}}} \\ 
			\cdashline{2-14}[3pt/2pt]
			~ & \multirow{2}*{F1} & Original & 77.38  & 90.94  & 73.44  & 67.12  & 75.31  & 89.95  & 67.52  & 78.59  & 86.81  & 79.96  & 78.70\textsubscript{$\pm$8.46}  \\ 
			~ & ~ & +UP4LS & 90.54  & 95.38  & 91.62  & 93.54  & 94.28  & 93.84  & 78.66  & 93.08  & 96.40  & 95.05  & \textbf{92.24\textsubscript{$\pm$5.08}} \textsubscript{\textcolor{red}{\textbf{($\uparrow$ 13.54)}}} \\  \midrule[0.7pt]
			
			\multirow{4}*{200:1} & \multirow{2}*{Acc} & Original & 74.31  & 85.75  & 68.80  & 68.25  & 72.71  & 75.37  & 62.56  & 71.64  & 75.16  & 75.98  & 73.05\textsubscript{$\pm$6.10}  \\ 
			~ &  & +UP4LS & 82.24  & 94.84  & 88.87  & 86.80  & 85.32  & 90.78  & 78.59  & 89.95  & 91.31  & 92.43  & \textbf{88.11\textsubscript{$\pm$4.94}} \textsubscript{\textcolor{red}{\textbf{($\uparrow$ 15.06)}}} \\ 
			\cdashline{2-14}[3pt/2pt]
			~ & \multirow{2}*{F1} & Original & 65.43  & 83.38  & 54.64  & 53.47  & 62.46  & 67.32  & 40.15  & 60.40  & 66.96  & 68.39  & 62.26\textsubscript{$\pm$11.40}  \\ 
			~ & ~ & +UP4LS & 78.83  & 94.74  & 87.50  & 84.80  & 82.43  & 89.87  & 76.85  & 88.85  & 90.20  & 91.82  & \textbf{86.59\textsubscript{$\pm$5.76}} \textsubscript{\textcolor{red}{\textbf{($\uparrow$ 24.33)}}} \\  \midrule[0.7pt]
			
			\multirow{4}*{500:1} & \multirow{2}*{Acc} & Original & 53.02  & 63.64  & 52.28  & 55.77  & 58.47  & 57.34  & 50.85  & 53.24  & 51.74  & 53.48  & 54.98\textsubscript{$\pm$3.91}  \\ 
			~ &  & +UP4LS & 80.54  & 88.20  & 84.43  & 86.80  & 80.35  & 76.00  & 64.29  & 85.48  & 83.01  & 89.27  & \textbf{81.84\textsubscript{$\pm$7.36}} \textsubscript{\textcolor{red}{\textbf{($\uparrow$ 26.86)}}} \\ 
			\cdashline{2-14}[3pt/2pt]
			~ & \multirow{2}*{F1} & Original & 11.38  & 42.86  & 8.73  & 20.70  & 28.97  & 25.60  & 3.33  & 12.16  & 6.74  & 13.02  & 17.35\textsubscript{$\pm$12.16}  \\ 
			~ & ~ & +UP4LS & 75.83  & 87.24  & 86.78  & 85.90  & 80.44  & 69.37  & 52.07  & 83.24  & 79.90  & 84.72  & \textbf{78.55\textsubscript{$\pm$10.83}} \textsubscript{\textcolor{red}{\textbf{($\uparrow$ 61.20)}}} \\  
			\toprule[1.3pt]

			\multicolumn{3}{c}{HiTIN \cite{HiTIN}} & U1 & U2 & U3 & U4 & U5 & U6 & U7 & U8 & U9 & U10 & Avg\textsubscript{$\pm$Std} \textsubscript{\textcolor{red}{\textbf{($\mathbf{\Delta}$Acc)}}} \\ \midrule[0.7pt]
			
			\multirow{4}*{50:1} & \multirow{2}*{Acc} & Original & 79.76  & 95.11  & 86.58  & 83.62  & 82.59  & 89.94  & 78.13  & 89.69  & 93.09  & 93.50  & 87.20\textsubscript{$\pm$5.99} \\ 
			~ &  & +UP4LS & 93.28  & 96.68  & 96.26  & 96.91  & 97.27  & 97.19  & 90.34  & 96.95  & 97.28  & 97.50  & \textbf{95.97\textsubscript{$\pm$2.32}} \textsubscript{\textcolor{red}{\textbf{($\uparrow$ 8.77)}}} \\
			\cdashline{2-14}[3pt/2pt]
			~ & \multirow{2}*{F1} & Original & 74.63  & 95.00  & 84.68  & 81.72  & 79.94  & 89.65  & 73.26  & 88.39  & 92.70  & 93.40  & 85.34\textsubscript{$\pm$7.77}  \\ 
			~ & ~ & +UP4LS & 92.88  & 96.60  & 96.15  & 96.81  & 97.20  & 97.09  & 89.90  & 96.89  & 97.20  & 97.44  & \textbf{95.82\textsubscript{$\pm$2.46}} \textsubscript{\textcolor{red}{\textbf{($\uparrow$ 10.48)}}} \\  \midrule[0.7pt]
			
			\multirow{4}*{100:1} & \multirow{2}*{Acc} & Original & 65.95  & 93.59  & 64.94  & 68.09  & 87.20  & 92.41  & 54.09  & 87.04  & 66.35  & 84.33  & 76.40\textsubscript{$\pm$13.96}  \\
			~ &  & +UP4LS & 89.05  & 95.80  & 92.43  & 94.90  & 94.00  & 94.72  & 81.04  & 93.95  & 94.77  & 96.07  & \textbf{92.67\textsubscript{$\pm$4.56}} \textsubscript{\textcolor{red}{\textbf{($\uparrow$ 16.27)}}} \\
			\cdashline{2-14}[3pt/2pt]
			~ & \multirow{2}*{F1} & Original & 48.51  & 92.82  & 46.28  & 55.22  & 88.05  & 90.78  & 17.26  & 84.87  & 53.93  & 76.16  & 65.39\textsubscript{$\pm$24.96}  \\ 
			~ & ~ & +UP4LS & 88.19  & 95.64  & 91.81  & 94.70  & 93.63  & 94.35  & 76.60  & 93.56  & 94.48  & 95.93  & \textbf{91.89\textsubscript{$\pm$5.82}} \textsubscript{\textcolor{red}{\textbf{($\uparrow$ 26.50)}}} \\  \midrule[0.7pt]
			
			\multirow{4}*{200:1} & \multirow{2}*{Acc} & Original & 55.14  & 90.73  & 63.01  & 67.51  & 70.86  & 85.13  & 52.93  & 79.54  & 53.96  & 90.14  & 70.90\textsubscript{$\pm$14.80}  \\ 
			~ &  & +UP4LS & 85.02  & 95.02  & 90.97  & 86.29  & 90.74  & 92.33  & 78.09  & 92.05  & 88.40  & 91.52  & \textbf{89.04\textsubscript{$\pm$4.86}} \textsubscript{\textcolor{red}{\textbf{($\uparrow$ 18.14)}}}\\ 
			\cdashline{2-14}[3pt/2pt]
			~ & \multirow{2}*{F1} & Original & 13.80  & 89.82  & 43.59  & 49.14  & 59.23  & 86.27  & 14.38  & 73.92  & 13.52  & 88.50  & 53.22\textsubscript{$\pm$31.43}  \\ 
			~ & ~ & +UP4LS & 82.33  & 94.92  & 90.63  & 84.33  & 89.70  & 91.71  & 69.82  & 89.28  & 86.56  & 90.75  & \textbf{87.00\textsubscript{$\pm$7.06}} \textsubscript{\textcolor{red}{\textbf{($\uparrow$ 33.78)}}} \\  \midrule[0.7pt]
			
			\multirow{4}*{500:1} & \multirow{2}*{Acc} & Original & 51.85  & 52.45  & 52.08  & 51.03  & 50.06  & 52.64  & 52.93  & 56.33  & 50.66  & 52.95  & 52.30\textsubscript{$\pm$1.72}  \\ 
			~ &  & +UP4LS & 81.72  & 86.88  & 83.75  & 84.99  & 84.21  & 75.04  & 72.71  & 87.95  & 83.66  & 88.21  & \textbf{82.91\textsubscript{$\pm$5.21}} \textsubscript{\textcolor{red}{\textbf{($\uparrow$ 30.61)}}} \\ 
			\cdashline{2-14}[3pt/2pt]
			~ & \multirow{2}*{F1} & Original & 7.95  & 9.33  & 9.72  & 4.44  & 0.94  & 8.15  & 9.72  & 35.41  & 1.75  & 12.31  & 9.97\textsubscript{$\pm$9.66}  \\ 
			~ & ~ & +UP4LS & 79.17  & 84.18  & 78.42  & 82.97  & 81.04  & 66.81  & 76.41  & 87.12  & 85.90  & 86.88  & \textbf{80.89\textsubscript{$\pm$6.17}} \textsubscript{\textcolor{red}{\textbf{($\uparrow$ 70.92)}}} \\  \bottomrule[1.3pt]
		\end{tabular}
	}
\end{table*}

\begin{table*}[!ht]
	\centering
	{\fontsize{8.5pt}{9.5pt}\selectfont
		\setlength{\tabcolsep}{1.55mm}
		\caption{Ablation experiment about the user features.}\label{a-a}
		\begin{tabular}{cccccccccccccc}
			\toprule[1.3pt]
			\multicolumn{3}{c}{\textbf{User features (\%)}} & U1 & U2 & U3 & U4 & U5 & U6 & U7 & U8 & U9 & U10 & Avg\textsubscript{$\pm$Std} \textsubscript{\textcolor{red}{\textbf{($\mathbf{\Delta}$Acc)}}} \\ \midrule[0.7pt]
			
			\multirow{4}*{50:1} &  \multirow{2}*{Acc} & Content & 89.46& 	94.67& 	90.59& 	90.52& 	91.57 &	95.79 &	80.27 &	91.99 &	92.56 &	95.95   & 91.34\textsubscript{$\pm$4.49}  \\	
			~ & & User+Content & 95.56  & 97.04  & 96.52  & 97.04  & 97.06  & 98.31  & 89.36  & 96.17  & 97.46  & 97.68  & \textbf{96.22\textsubscript{$\pm$2.53}} \textsubscript{\textcolor{red}{\textbf{($\uparrow$ 4.88)}}} \\
			\cdashline{2-14}[3pt/2pt]
			~ & \multirow{2}*{F1}& Content & 88.28 &	94.37 &	89.41 &	89.47 &	90.76 &	95.58 &	74.87 &	91.23 &	91.91 &	95.79 & 90.17\textsubscript{$\pm$5.98}  \\ 
			~ & ~ & User+Content & 95.51  & 96.99  & 96.42  & 97.01  & 97.01  & 98.31  & 88.32  & 96.06  & 97.41  & 97.65  & \textbf{96.07\textsubscript{$\pm$2.84}} \textsubscript{\textcolor{red}{\textbf{($\uparrow$ 5.90)}}} \\  \midrule[0.7pt]
			
			\multirow{4}*{100:1} & \multirow{2}*{Acc} & Content & 78.17 &	91.61 &	80.54 &	79.45 &	79.76 &	92.01 &	67.83 &	79.76 &	87.39 &	87.32 &	82.38\textsubscript{$\pm$7.32}  \\
			~ &  & User+Content & 93.19  & 95.93  & 93.98  & 95.13  & 94.28  & 95.00  & 82.40  & 94.54  & 94.67  & 95.23  & \textbf{93.44\textsubscript{$\pm$3.95}} \textsubscript{\textcolor{red}{\textbf{($\uparrow$ 11.06)}}} \\
			\cdashline{2-14}[3pt/2pt]
			~ & \multirow{2}*{F1} & Content & 72.02 &	90.84 &	75.83 &	74.00 &	74.10 &	91.32 &	52.50 &	74.39 &	85.38 &	85.38 &	77.58\textsubscript{$\pm$11.47}  \\ 
			~ & ~ & User+Content & 93.00  & 95.59  & 93.57  & 94.93  & 93.73  & 94.74  & 79.80  & 94.44  & 94.39  & 94.99  & \textbf{92.92\textsubscript{$\pm$4.67}} \textsubscript{\textcolor{red}{\textbf{($\uparrow$ 15.34)}}} \\  \midrule[0.7pt]
			
			\multirow{4}*{200:1} & \multirow{2}*{Acc} & Content & 58.08  &	79.14  &	67.46  &	72.68  &	71.06 & 	84.91 & 	60.70  &	71.52  &	83.70  &	84.52  &	73.38\textsubscript{$\pm$9.64}  \\
			~ &  & User+Content & 84.61  & 94.32  & 92.49  & 88.19  & 91.02  & 91.79  & 77.47  & 90.95  & 89.28  & 92.03  & \textbf{89.22\textsubscript{$\pm$4.92}} \textsubscript{\textcolor{red}{\textbf{($\uparrow$ 15.84)}}} \\
			\cdashline{2-14}[3pt/2pt]
			~ & \multirow{2}*{F1} & Content & 26.37 &	73.73& 	51.57& 	62.41& 	59.27& 	82.16 &	32.38& 	60.11& 	80.23& 	81.60& 	60.98\textsubscript{$\pm$19.77}  \\ 
			~ & ~ & User+Content & 82.52  & 94.20  & 92.03  & 86.24  & 90.07  & 90.81  & 71.06  & 89.71  & 87.98  & 91.32  & \textbf{87.59\textsubscript{$\pm$6.66}} \textsubscript{\textcolor{red}{\textbf{($\uparrow$ 26.61)}}} \\  \midrule[0.7pt]
			
			\multirow{4}*{500:1} & \multirow{2}*{Acc} & Content & 52.18 &	62.27 &	54.42& 	53.10 &	58.00 &	69.06 &	51.57 &	58.09 &	53.59 &	60.89 &	57.32\textsubscript{$\pm$5.54}  \\ 
			~ &  & User+Content & 82.48  & 88.70  & 83.33  & 87.80  & 81.75  & 80.49  & 71.97  & 88.77  & 79.70  & 88.89  & \textbf{83.39\textsubscript{$\pm$5.40}} \textsubscript{\textcolor{red}{\textbf{($\uparrow$ 26.07)}}} \\ 
			\cdashline{2-14}[3pt/2pt]
			~ & \multirow{2}*{F1} & Content & 8.60 &	38.77 &	14.70 &	11.16 &	26.29 &	54.84 &	6.11 &	25.53 &	13.13& 	35.39 &	23.45\textsubscript{$\pm$15.74}  \\ 
			~ & ~ & User+Content & 79.36  & 86.51  & 80.20  & 86.69  & 77.62  & 75.40  & 62.91  & 87.48  & 72.89  & 87.71  & \textbf{79.68\textsubscript{$\pm$7.97}} \textsubscript{\textcolor{red}{\textbf{($\uparrow$ 56.23)}}} \\  \bottomrule[1.3pt]
		\end{tabular}
	}
\end{table*}

\begin{table*}[!htbp]
	\centering
	{\fontsize{8.5pt}{10.5pt}\selectfont
		\setlength{\tabcolsep}{0.3mm}
		\caption{Ablation experiment about the attention fusion.}\label{a-b}
		\begin{tabular}{ccccccccccccc}
			\toprule[1.3pt]
			\multicolumn{2}{c}{\textbf{Fusion (\%)}}   & U1    & U2    & U3    & U4    & U5    & U6    & U7    & U8    & U9    & U10   & Avg\textsubscript{$\pm$Std} \\
			\midrule[0.6pt]
			\multirow{2}*{Acc} & Concat   & 95.17\textsubscript{$\pm$1.53} & 96.85\textsubscript{$\pm$1.48} & 96.84\textsubscript{$\pm$1.35} & 96.84\textsubscript{$\pm$0.87} & 95.79\textsubscript{$\pm$0.79} & 97.70\textsubscript{$\pm$0.60} & 90.41\textsubscript{$\pm$1.95} & 95.55\textsubscript{$\pm$2.19} & 97.49\textsubscript{$\pm$0.86} & 96.14\textsubscript{$\pm$1.00} & 95.88\textsubscript{$\pm$2.09} \\
			& Attn   & 96.9 & 96.5 & 97.72 & 96.7 & 97.76 & 98.86 & 91.67 & 95.66 & 97.93 & 97.86 & \textbf{96.76}\textsubscript{$\pm$2.00} \\
			\cdashline{1-13}[3pt/2pt]
			
			\multirow{2}*{F1}& Concat    & 94.68\textsubscript{$\pm$1.97} & 96.76\textsubscript{$\pm$1.54} & 96.55\textsubscript{$\pm$1.51} & 96.79\textsubscript{$\pm$0.96} & 95.61\textsubscript{$\pm$0.86} & 97.65\textsubscript{$\pm$0.64} & 88.80\textsubscript{$\pm$2.72} & 95.30\textsubscript{$\pm$2.44} & 97.43\textsubscript{$\pm$0.92} & 95.98\textsubscript{$\pm$1.10} & 95.56\textsubscript{$\pm$2.55} \\
			& Attn    & 96.9 & 96.43 & 97.68 & 96.6 & 97.71 & 98.85 & 91.28 & 95.51 & 97.90 & 97.81 & \textbf{96.67}\textsubscript{$\pm$2.11} \\
			\bottomrule[1.3pt]
		\end{tabular}%
	}
\end{table*}%

\begin{table*}[!htbp]
	\centering
	{\fontsize{8.5pt}{10.5pt}\selectfont
		\setlength{\tabcolsep}{2.2mm}
		\caption{Significance test in the comparison experiment.}\label{stest}
		\begin{tabular}{cccccccccc}
			\toprule[1.3pt]
			\multicolumn{2}{c}{\multirow{2}{*}{\textbf{Significance test}}} & \multicolumn{4}{c}{FETS} & \multicolumn{4}{c}{TS\_RNN} \\
			\multicolumn{2}{c}{} & 50:1   & 100:1   & 200:1   & 500:1   & 50:1   & 100:1   & 200:1   & 500:1   \\
			\midrule[0.6pt]
			\multirow{2}{*}{T} & Acc & 4.17E-20 & 2.51E-17 & 1.55E-14 & 3.06E-11 & 9.30E-24 & 1.22E-17 & 8.76E-16 & 8.46E-13 \\
			& F1  & 2.02E-24 & 1.09E-21 & 2.02E-17 & 2.08E-14 & 1.05E-28 & 1.30E-22 & 8.75E-20 & 1.04E-15 \\
			\midrule[0.6pt]
			\multirow{2}{*}{Mann–Whitney U} & Acc & 0.000172 & 8.74E-05 & 8.74E-05 & 8.74E-05 & 0.000177 & 0.000149 & 0.000110 & 8.74E-05 \\
			& F1  & 0.000172 & 8.74E-05 & 8.74E-05 & 8.74E-05 & 0.000178 & 0.000149 & 0.000110 & 8.74E-05 \\
			\toprule[1.3pt]
			
			\multicolumn{2}{c}{\multirow{2}{*}{\textbf{Significance test}}} & \multicolumn{4}{c}{Zou} & \multicolumn{4}{c}{SSLS}  \\
			\multicolumn{2}{c}{} & 50:1   & 100:1   & 200:1   & 500:1   & 50:1   & 100:1   & 200:1   & 500:1   \\
			\midrule[0.6pt]
			\multirow{2}[0]{*}{T} & Acc & 0.009348 & 0.000152 & 4.30E-05 & 4.59E-08 & 0.005720 & 0.001382 & 0.001477 & 5.89E-10 \\
			& F1  & 0.015220 & 0.000506 & 0.000492 & 3.14E-07 & 0.009517 & 0.002618 & 0.004584 & 2.05E-09 \\
			\midrule[0.6pt]
			\multirow{2}{*}{Mann–Whitney U} & Acc & 0.002202 & 0.000582 & 0.000329 & 0.000246 & 0.001314 & 0.001007 & 0.001314 & 0.000182 \\
			& F1  & 0.001699 & 0.000439 & 0.000329 & 0.000246 & 0.001007 & 0.001007 & 0.001314 & 0.000182 \\
			\toprule[1.3pt]
			
			\multicolumn{2}{c}{\multirow{2}{*}{\textbf{Significance test}}} & \multicolumn{4}{c}{LSFLS} & \multicolumn{4}{c}{HypEmo} \\
			\multicolumn{2}{c}{} & 50:1   & 100:1   & 200:1   & 500:1   & 50:1   & 100:1   & 200:1   & 500:1   \\
			\midrule[0.6pt]
			\multirow{2}{*}{T} & Acc & 0.045666 & 0.001358 & 0.000163 & 9.79E-08 & 0.000673 & 0.000402 & 9.81E-06 & 6.67E-09 \\
			& F1  & 0.045557 & 0.001811 & 0.000350 & 2.41E-07 & 0.000638 & 0.000394 & 1.07E-05 & 5.87E-10 \\
			\midrule[0.6pt]
			\multirow{2}{*}{Mann–Whitney U} & Acc & 0.025748 & 0.000768 & 0.001706 & 0.000182 & 0.002827 & 0.001007 & 0.000439 & 0.000182 \\
			& F1  & 0.025748 & 0.000768 & 0.001706 & 0.000182 & 0.002827 & 0.000768 & 0.000439 & 0.000182 \\
			\toprule[1.3pt]
			
			\multicolumn{2}{c}{\multirow{2}{*}{\textbf{Significance test}}} & \multicolumn{4}{c}{HiTIN} &    /   &    /   &   /    & / \\
			\multicolumn{2}{c}{} & 50:1   & 100:1   & 200:1   & 500:1   &  /  &      / &     /  &  /\\
			\midrule[0.6pt]
			\multirow{2}[0]{*}{T} & Acc & 0.000416 & 0.002525 & 0.001701 & 8.22E-13 &     /  &    /   &   /    &  /\\
			& F1  & 0.000727 & 0.004255 & 0.003835 & 1.41E-13 &     /  &     /  &     /  &  /\\
			\midrule[0.6pt]
			\multirow{2}[1]{*}{Mann–Whitney U} & Acc & 0.000768 & 0.001706 & 0.003610 & 0.000182 &   /    &    /   &  /     & / \\
			& F1  & 0.000765 & 0.001314 & 0.007284 & 0.000181 &    /   &   /    &    /   &  /\\
			\bottomrule[1.3pt]  
		\end{tabular}%
	}
\end{table*}%

\end{document}